\documentclass[10pt,journal,compsoc]{IEEEtran}

\ifCLASSOPTIONcompsoc
  \usepackage[nocompress]{cite}
\else
  \usepackage{cite}
\fi
\usepackage{ifthen}  
\usepackage{subcaption} 
\usepackage{amsmath}
\usepackage{mathtools}
\usepackage{amssymb}
\usepackage[dvipsnames]{xcolor}  
\usepackage{array}  
\usepackage[hyphens]{url}
\usepackage[hidelinks]{hyperref}
\usepackage{multirow}
\usepackage{pdfpages}
\hypersetup{breaklinks=true}
\newcommand{\colornum}[1]{{\textbf{\color[RGB]{39, 158, 206}#1}}}
\newcommand{\mat}[1]{\MakeUppercase{\mathbf{#1}}}
\renewcommand{\vec}[1]{\MakeLowercase{\mathbf{#1}}}
\newcommand{\tmat}[1]{$\mat{#1}$}

\newcommand{\myparagraph}[2][0]{\vspace{0em}\noindent{\bf #2}}

\begin{document}
\title{Optimising for Interpretability:\\ Convolutional Dynamic Alignment Networks}

\author{Moritz~Böhle,
        Mario~Fritz, 
        and~Bernt~Schiele%
\IEEEcompsocitemizethanks{\IEEEcompsocthanksitem Moritz Böhle is the corresponding author. \protect\\
He is with the Department of Computer Vision and Machine Learning, Max Planck Institute for Informatics, Saarbrücken 66123, Germany. E-mail: mboehle@mpi-inf.mpg.de 
\protect\\
\IEEEcompsocthanksitem Bernt Schiele is with the Department of Computer Vision and Machine Learning, Max Planck Institute for Informatics, Saarbrücken 66123, Germany. E-mail: schiele@mpi-inf.mpg.de
\protect\\
\IEEEcompsocthanksitem
Mario Fritz is with the CISPA Helmholtz Center for Information Security, Saarbrücken 66123, Germany. E-mail: fritz@cispa.de.
}
\thanks{
}
}

\markboth{
Published in IEEE Transactions on Pattern Analysis and Machine Intelligence (Volume 45, Issue: 6, 01 June 2023).%
}%
{
}
%



\IEEEtitleabstractindextext{%
\begin{abstract}
We introduce a new family of neural network models called Convolutional Dynamic Alignment Networks (CoDA Nets),
which are performant classifiers with a high degree of inherent interpretability. 
Their core building blocks are Dynamic Alignment Units (DAUs), which are optimised to transform their inputs with dynamically computed weight vectors that align with task-relevant patterns. As a result, CoDA Nets model the classification prediction through a series of input-dependent linear transformations, allowing for linear decomposition of the output into individual input contributions.
    Given the alignment of the DAUs, the resulting contribution maps 
    align with discriminative input patterns. 
These model-inherent decompositions are of high visual quality and  outperform existing attribution methods under quantitative metrics. 
Further, CoDA Nets constitute performant classifiers, achieving on par results 
to ResNet and VGG models on e.g. CIFAR-10 and TinyImagenet. Lastly, CoDA Nets can be combined with conventional neural network models to yield powerful classifiers that more easily scale to complex datasets such as Imagenet whilst exhibiting an increased \emph{interpretable depth}, i.e., the output can be explained well in terms of contributions from intermediate layers within the network.
\end{abstract}

\begin{IEEEkeywords}
Explainability in Deep Learning, Convolutional Neural Networks
\end{IEEEkeywords}
}

\maketitle
\IEEEdisplaynontitleabstractindextext
\IEEEpeerreviewmaketitle

\IEEEraisesectionheading{%
\section{Introduction}\label{sec:introduction}}
\begin{figure*}[!ht]
    \centering
    \begin{subfigure}[b]{.975\textwidth}
    \centering
    \includegraphics[width=1\textwidth]{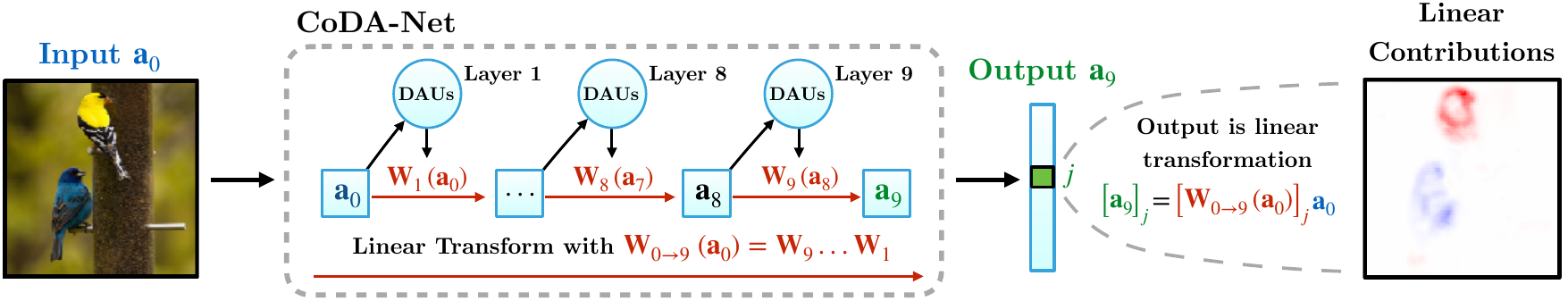}
    \end{subfigure}
    \caption{\small {
    Sketch of a 9-layer CoDA-Net, which computes its {\bf \color[RGB]{0, 136, 43} output $\vec {a_9}$} for an  {\bf \color[RGB]{3, 101, 192}input $\vec {a_0}$} as a linear transform via a  matrix {\color[RGB]{200, 37, 6}$\mat {W_{0\rightarrow9}(\vec a_0)}$}. As such, the output can be linearly decomposed into input contributions (see right). This 'global' transformation matrix {\color[RGB]{200, 37, 6}$\mat {w_{0\rightarrow9}}$} is computed successively via multiple layers of Dynamic Alignment Units (DAUs). These layers, in turn, produce intermediate linear transformation matrices {\color[RGB]{200, 37, 6}$\mat w_l(\vec a_{l-1})$} that align with the inputs of layer $l$. As a result, the combined matrix {\color[RGB]{200, 37, 6}$\mat {w_{0\rightarrow9}}$} also aligns well with task-relevant patterns. Positive (negative) contributions for the class `goldfinch' are shown in red (blue).} 
    }
    \label{fig:teaser}
\end{figure*}
\IEEEPARstart{N}{eural} networks are powerful models that excel at a wide range of tasks.
However, they are notoriously difficult to interpret and extracting explanations 
    for their predictions is an open research problem. Linear models, in contrast, are generally considered interpretable, because
    the \emph{contribution} 
    (`the weighted input') of every dimension to the output is explicitly given.
Interestingly, many modern neural networks implicitly model the output as a linear transformation of the input; a ReLU-based~\cite{nair2010rectified} neural network, e.g.,
    is piece-wise linear and the output thus a linear transformation of the input, cf.~\cite{montufar2014number}.
    However, due to the highly non-linear manner in which these linear transformations are `chosen', the corresponding contributions per input dimension do not seem to represent the learnt model parameters well, cf.~\cite{adebayo2018sanity}, and a lot of research is being conducted to find better explanations for the decisions of such neural networks, cf.~\cite{simonyan2013deep,springenberg2014striving,zhou2016CAM,selvaraju2017grad,shrikumar2017deeplift,sundararajan2017axiomatic,srinivas2019full,bach2015pixel}.
    
In this work, we introduce a novel network architecture, the \textbf{Convolutional Dynamic Alignment Networks (CoDA Nets)}, {for which the model-inherent contribution maps are faithful projections of the internal computations and thus good `explanations' of the model prediction.} 
There are two main components to the interpretability of the CoDA Nets. 
    First, the CoDA Nets are \textbf{dynamic linear}, i.e., they compute their outputs through a series of input-dependent linear transforms, which are based on our novel \mbox{\textbf{Dynamic Alignment Units (DAUs)}}.  
    As in linear models, the output can thus be decomposed into individual input contributions, see Fig.~\ref{fig:teaser}.
    Second, the DAUs  
    are structurally biased to compute weight vectors that \textbf{align with \mbox{relevant} patterns} in their inputs. 
In combination, the CoDA Nets thus inherently produce contribution maps that are `optimised for interpretability': 
since each linear transformation matrix and thus their combination is optimised to align with discriminative features, the contribution maps reflect the most discriminative features \emph{as used by the model}.

With this work, we present a new direction for building inherently more interpretable neural network architectures with high modelling capacity.
In detail, we would like to highlight the following contributions:
\begin{enumerate}
    \item We introduce the concept of Dynamic Alignment Units (DAUs), which improve the interpretability of neural networks and have two key properties:
    they are 
    \emph{dynamic linear}
    and need to align their dynamically computed weights their inputs to achieve large outputs,
    since the norm of their weights is explicitly constrained.
    \item 
    Apart from normalising the dynamically computed DAU weights by their vector norm, we show that similar results can be achieved when normalising them by an \emph{upper bound} of their norm which results in more efficient DAUs.
    
    \item We show that networks of DAUs \emph{inherit} the dynamic linearity and the alignment properties from their constituent DAUs. In particular, we introduce Convolutional Dynamic Alignment Networks (CoDA Nets), which are built out of multiple layers of DAUs. As a result, the \emph{model-inherent contribution maps} of CoDA Nets highlight discriminative patterns in the input.
    \item We further show that the alignment of the DAUs can be promoted
    by applying a `temperature scaling' to the final output of the CoDA Nets.
    \item We show that the resulting contribution maps perform well under commonly employed \emph{quantitative} criteria for attribution methods. Moreover, under \emph{qualitative} inspection, we note that they exhibit a high degree of detail.
    \item 
    We analyse how the models are affected by different normalisation functions in the DAU weight calculation in terms of accuracy, interpretability, as well as efficiency.
    
    \item Beyond interpretability,  
    CoDA Nets are performant classifiers and yield competitive classification accuracies on the CIFAR-10 and TinyImagenet datasets.
    \item We show that CoDA Nets can be seamlessly combined with conventional networks. The resulting hybrid networks exhibit an increased 'interpretable depth' whilst taking advantage of the efficiency and strong modelling capacity of the base networks. Such networks hold great potential for designing models that are inherently interpretable up to a user-defined minimal  resolution.
\end{enumerate}

\section{Related Work}
\label{sec:related_work}
\myparagraph[0]{Interpretability.}
In order to make machine learning models more interpretable, a variety of techniques has been developed. While there are different ways in which interpretability can be defined, in this work we focus on attributing importance values to input features; for an extensive review regarding interpretability in machine learning, see \cite{linardatos2021explainable}.

Current techniques for interpreting neural networks via importance attribution can broadly be split into two categories: deriving \emph{post-hoc} explanations and developing inherently interpretable models. In the following, we will first describe common approaches for post-hoc interpretability.

On the one hand, 
    research regarding post-hoc explanations has been undertaken to develop model-agnostic explanation methods for which the model behaviour
    under perturbed inputs is analysed; this includes among others \cite{lundberg2017unified,petsiuk2018rise,lime}.
    While their generality and the applicability to any model are advantageous,
    these methods typically require evaluating the respective model several times and are therefore costly
    approximations of model behaviour. 
    Further, they rely on the assumption that the models generalise to such out-of-distribution (OOD) data (the perturbed / occluded inputs) in a stable manner, such that the outputs on the OOD data can be used to explain model behaviour on in-distribution data. 

On the other hand,
    many techniques that explicitly take advantage of the internal computations have been proposed for explaining
    the model predictions, including, for example, \cite{simonyan2013deep,springenberg2014striving,zhou2016CAM,selvaraju2017grad,shrikumar2017deeplift,sundararajan2017axiomatic,srinivas2019full,bach2015pixel}. Such methods typically distribute importance values layer by layer in a backward pass and introduce different rules for this re-distribution. Similarly, the model-inherent contribution maps in the CoDA Nets can also be obtained as a layer-wise decomposition of the output. However, there is a key difference: producing well-aligned linear decompositions is actually optimised for in the CoDA Nets during training, whereas other methods are developed without taking the optimisation procedure into account. As a result, the \emph{inherent} explanations in the CoDA Nets lend themselves better to understanding the model outputs.  \\
In contrast to techniques that aim to explain models \emph{post-hoc},
some recent work has focused on designing new types of network architectures, which are \emph{inherently} more interpretable.
Examples of this are the prototype-based neural networks~\cite{chen2019looks}, the BagNet~\cite{brendel2018approximating}
and the self-explaining neural networks (SENNs)~\cite{melis2018towards}.
Similarly to our proposed architectures,
    the SENNs and the BagNets derive their explanations 
    from a linear decomposition of the output into contributions from the input (features).
This \emph{dynamic linearity}, i.e., the property that the output is computed via some form of an input-dependent linear mapping, is additionally shared by the entire model family of piece-wise linear networks (e.g., ReLU-based networks). In fact, the contribution maps of the CoDA Nets are conceptually similar to  evaluating the `Input$\times$Gradient' (IxG), cf.~\cite{adebayo2018sanity}, on piece-wise linear models, which also yields a linear decomposition in form of a contribution map.
However, in contrast to the piece-wise linear functions, we combine this \emph{dynamic linearity} with a structural bias towards an alignment between the contribution maps and the discriminative patterns in the input. This results in explanations of much higher quality, whereas IxG on piece-wise linear models has been found to yield unsatisfactory explanations of model behaviour~\cite{adebayo2018sanity}.

\myparagraph{Architectural similarities.} In our CoDA Nets, the convolutional kernels are dependent on the specific patch that they are applied on; i.e., a CoDA layer might apply different filters at every position in the input. As such, these layers can be regarded as an instance of dynamic local filtering layers as introduced in~\cite{jia2016dynamic}.
Further, our dynamic alignment units (DAUs) share some high-level similarities to attention networks, cf.~\cite{xu2015show}, in the sense that each DAU has a limited budget to distribute over its dynamic weight vectors (bounded norm), which is then used to compute a weighted sum. However, whereas in attention networks the weighted sum is typically computed over vectors (the 'value vectors') which differ from the input to the attention module (the 'key' and 'query' vectors), a DAU outputs a \emph{scalar} which is a weighted sum of all scalar entries in the input. Moreover, we note that at their optimum (maximal average output over a set of inputs), the DAUs solve a constrained low-rank matrix approximation problem~\cite{eckart1936approximation}. While low-rank approximations have been used for increasing parameter efficiency in neural networks, cf.~\cite{yu2017compressing}, this concept has to the best of our knowledge not been used in order to endow neural networks with a structural bias towards finding low-rank approximations of the input for increased interpretability in classification tasks. {Lastly, the CoDA Nets 
are related to capsule networks. However, whereas in classical capsule networks the activation vectors of the capsules directly serve as input to the next layer, in CoDA Nets the corresponding vectors are used as convolutional filters. 
We include a detailed comparison in the supplement. }

\section{Dynamic Alignment Networks}
\label{sec:DANets}
In this section, we present our novel type of network architecture: the Convolutional Dynamic Alignment Networks (CoDA Nets). For this, we first introduce Dynamic Alignment Units (DAUs) as the basic building blocks of CoDA Nets and discuss two of their key properties in sec.~\ref{subsec:align_units}. Concretely, we show that these units linearly transform their inputs with dynamic (input-dependent) weight vectors and, additionally, that they are biased to align these weights with the input during optimisation. 
Given the computational costs of evaluating DAUs, in sec.~\ref{subsec:eDAUs} we further present an alternative formulation of the DAUs for increased efficiency.
We then discuss how DAUs can be used for classification (sec.~\ref{subsec:classification}) and how we build performant networks out of multiple layers of convolutional DAUs (sec.~\ref{subsec:coda}). Importantly, the resulting \emph{linear decompositions} of the network outputs are optimised to align with discriminative patterns in the input, making them highly suitable for interpreting the network predictions. 

\begin{figure}[t!]
    \centering
    \hspace{-.25em}
    \begin{subfigure}[b]{0.48\textwidth}
    \includegraphics[width=\textwidth]{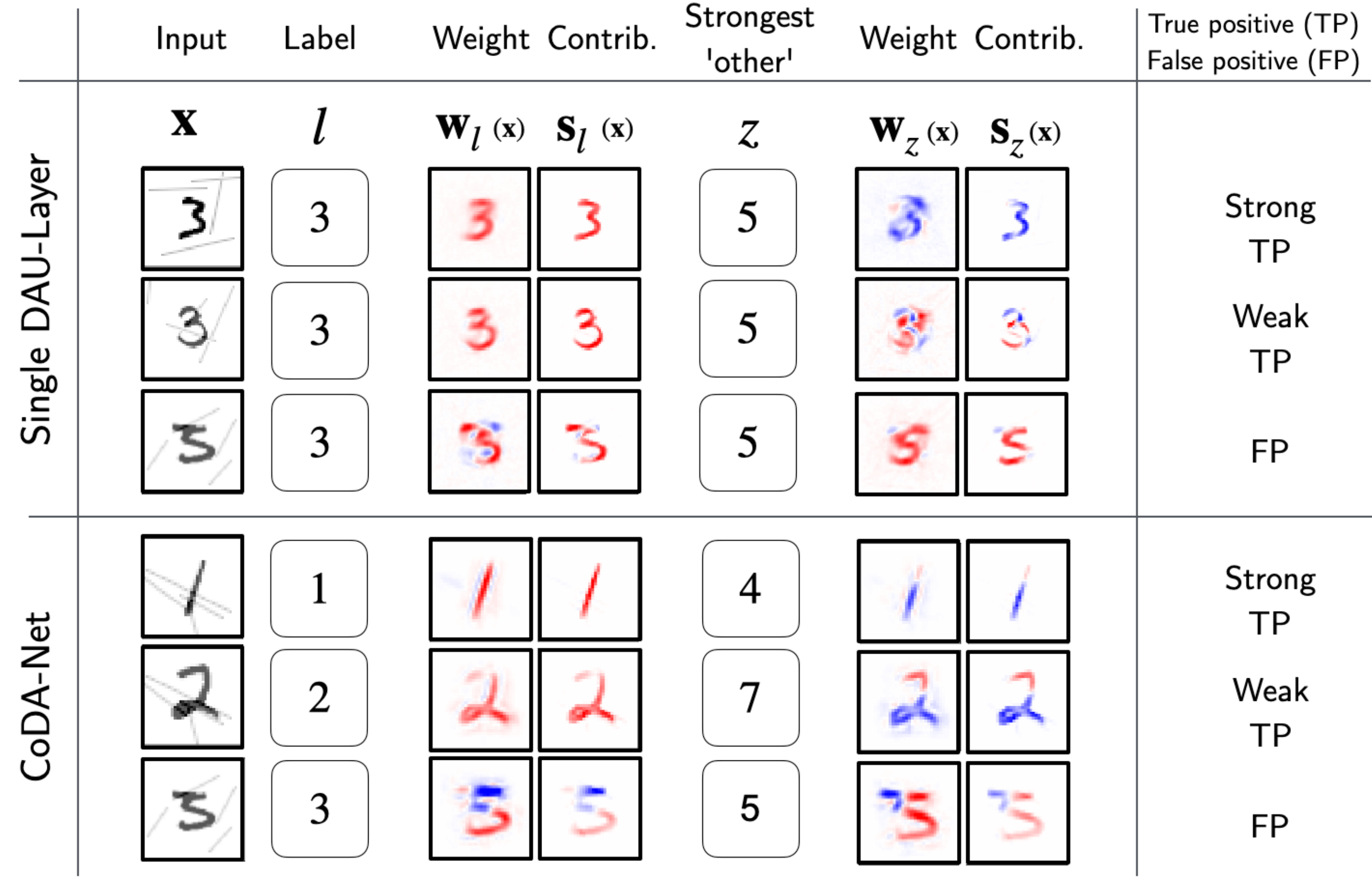}
    \end{subfigure}
    \caption{\small 
    For different inputs $\vec x$, we visualise the linear weights and contributions (for the single layer, see eq.~\eqref{eq:contrib_1}, for the CoDA-Net eq.~\eqref{eq:contrib}) for the ground truth label $l$ and the strongest non-label output $z$. 
    As can be seen, the weights align well with the input images.
    The first three rows are based on a single DAU layer, the last three on a 5 layer CoDA-Net. The first two samples (rows) per model are correctly classified and the last one is misclassified. 
    }
    \label{fig:alignment}
\end{figure}

In particular, we structure this section around the following \textbf{three important properties} (\colornum{P1-P3}) of the DAUs:
\\[.25em]
\colornum{P1: Dynamic linearity.} The DAU output $o$ is computed as a dynamic (input-dependent) linear transformation of the input $\vec x$, such that \mbox{$o=\vec w(\vec x)^T\vec x=\sum_jw_j(\vec x)x_j$}. 
Hence, 
$o$ can be decomposed into contributions  
from individual input dimensions, which are given by $w_j(\vec x)x_j$ for dimension $j$.
\\[.5em]
\colornum{P2: Alignment maximisation.} Maximising the average output of a single DAU over a set of inputs $\vec x_i$ 
maximises the alignment between inputs $\vec x_i$ and the weight vectors $\vec w(\vec x_i)$. As the modelling capacity of $\vec w(\vec x)$ is restricted, $\vec w(\vec x)$ will encode the most frequent patterns in the set of inputs $\vec x_i$.
\\[.5em]
\colornum{P3: Inheritance.} When combining multiple DAU layers to form a \mbox{Dynamic} Alignment Network (DA Net), the properties \colornum{P1} and \colornum{P2} are \emph{inherited}: DA Nets are dynamic linear (\colornum{P1}) and maximising the last layer's output induces an output maximisation in the constituent DAUs (\colornum{P2}).
\\[.5em]
These properties increase the interpretability
of a DA Net, such as a CoDA Net (sec.~\ref{subsec:coda}) for the following reasons.
First, the output of a DA Net can be decomposed into contributions from the individual input dimensions, similar to linear models (cf.~Fig.~\ref{fig:teaser}, \colornum{P1} and \colornum{P3}).
Second, we note that optimising a neural network for classification applies a maximisation to the outputs of the last layer for every sample. 
This maximisation aligns the dynamic weight vectors $\vec w(\vec x)$ of the constituent DAUs of the DA Net with their respective inputs (cf.~Fig.~\ref{fig:alignment} as well as \colornum{P2} and \colornum{P3}).

 Importantly, the weight vectors will align with the \emph{discriminative} patterns in their inputs when optimised for classification as we show in sec.~\ref{subsec:classification}.
As a result, the model-inherent contribution maps of CoDA Nets are optimised to align well with \emph{discriminative input patterns} in the input image 
{and  the interpretability of our models thus forms part of the global optimisation procedure.}\\[-1em]
\subsection{Dynamic Alignment Units}
\label{subsec:align_units}
We define the Dynamic Alignment Units (DAUs) by
\begin{align}
    \label{eq:au}
    \text{DAU}(\vec x) = g(\mat a \mat b\vec x +\vec b)^T \vec x = \vec w(\vec x)^T\, \vec x\quad \textbf{.}
\end{align}
Here, $\vec x\in\mathbb R^{d}$ is an input vector, $\mat a\in\mathbb R^{d\times r}$ and $\mat b \in \mathbb R^{r\times d}$ are trainable transformation matrices, $\vec b\in\mathbb R^{d}$ a trainable bias vector, and \mbox{$g(\vec u)=\alpha(||\vec u||)\vec u$} is a non-linear function that scales the norm of its input. {In contrast to using a single matrix $\mat m \in\mathbb R^{d\times d}$, using $\mat{ab}$ allows us to control the maximum rank $r$ of the transformation and to reduce the number of parameters}; we will hence refer to $r$ as the rank of a DAU. 
As can be seen by the right-hand side of eq.~\eqref{eq:au}, the DAU linearly transforms the input $\vec x$ (\colornum{P1}). {At the same time, given the quadratic form ($\vec x^T\mat B^T\mat A^T\vec x$) and the  rescaling function $\alpha(||\vec u||)$, the output of the DAU is a non-linear function of its input. 
In the context of DAUs, we are particularly interested in functions that constrain the norm of the weight vectors $\vec w(\vec x)$, such as, e.g., rescaling to unit norm ($\text{L2}$) or the squashing function ($\text{SQ}$, see \cite{sabour2017dynamic}):}
\begin{align}
    \label{eq:nonlin}
    \text{L2}(\vec u) = \frac{\vec u}{||\vec u||_2} \;\;\text{and}\;\;
    \text{SQ}(\vec u) = \text{L2}(\vec u) \times \frac{||\vec u||^2_2}{1+||\vec u||_2^2}
\end{align}
In sec.~\ref{subsec:eDAUs}, we further present an approximation to these rescaling functions, which lowers the computational cost of the DAUs whilst maintaining their bounding property $||\vec w(\vec x)|| \leq 1$. Given such a bound on $\vec w( \vec x)$, the output of the DAUs will be upper-bounded by the norm of the input:
{\begin{align}
    \text{DAU}(\vec x) = 
    ||\vec w(\vec x)|| \hspace{.2em} ||\vec x|| \cos(\angle(\vec x, \vec w(\vec x)))\leq ||\vec x||
    \label{eq:bound}
\end{align}}
As a corollary, for a given input $\vec x_i$, the DAUs can only achieve this upper bound if $\vec x_i$ is an eigenvector (EV) of the linear transform $\mat{AB}\vec x+ \vec b$. Otherwise, the cosine in eq.~\eqref{eq:bound} will not be maximal\footnote{
Note that $\vec w(\vec x)$ is proportional to $\mat{ab}\vec x + \vec b$. The cosine in eq.~\eqref{eq:bound}, in turn, is maximal if and only if $\vec w(\vec x_i)$ is proportional to $\vec x_i$ and thus, by transitivity, if $\vec x_i$ is proportional to $\mat{ab}\vec x_i + \vec b$. This means that $\vec x_i$ has to be an EV of $\mat{ab}\vec x +\vec b$ to achieve maximal output.}. 
As can be seen in eq.~\eqref{eq:bound}, maximising the average output of a DAU over a set of inputs $\{\vec x_i|\,i=1, ..., n\}$
maximises the alignment between $\vec w(\vec x)$ and $\vec x$ (\colornum{P2}).
In particular, it optimises the parameters of the DAU such that the \emph{most frequent input patterns} are encoded as EVs in the linear transform $\mat{ab}\vec x + \vec b$, similar to an $r$-dimensional PCA decomposition ($r$ the rank of $\mat{ab}$). In fact, as discussed in the supplement, the optimum of the DAU maximisation solves a low-rank matrix approximation~\cite{eckart1936approximation} problem similar to singular value decomposition.
\begin{figure}[t!]
    \centering
    \includegraphics[height=6.5em]{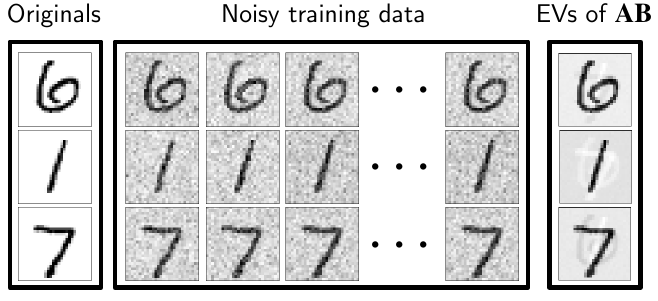}
    \caption{\small Eigenvectors (EVs) of \tmat{AB} after maximising the output of a rank-3 DAU over a set of noisy samples of 3 MNIST digits. Effectively, the DAUs encode the most frequent components in their EVs, similar to a principal component analysis (PCA).
    }
    \label{fig:EVs}
\end{figure}
As an illustration of this property, in Fig.~\ref{fig:EVs} we show the 3 EVs\footnote{Given $r=3$, the EVs maximally span a 3-dimensional subspace.} of matrix $\mat{ab}$ (with rank $r=3$, bias $\vec b=\vec 0$) after optimising a DAU over a set of $n$ noisy samples of 3 specific MNIST~\cite{lecun2010mnist} images; for this, we used $n=3072$ and zero-mean Gaussian noise. As expected, the EVs of \tmat{ab} encode the original, noise-free images, since this on average maximises the alignment (eq.~\eqref{eq:bound}) between the weight vectors $\vec w(\vec x_i)$ and the input samples $\vec x_i$ over the dataset.

\subsection{Efficient DAUs: Bounding the Bound}
\label{subsec:eDAUs}

As discussed in the previous section, we introduce a norm constraint for the DAU weights $\vec w(\vec x)$ to ensure that large outputs can only be achieved for well-aligned weights. However, the explicit norm constraint on $\vec w(\vec x)$ requires its explicit calculation, which we have observed to significantly impact the evaluation time of DAUs. Therefore, we evaluate an additional formulation of the DAUs in which we only \emph{constrain an upper bound} of the norm of $\vec w(\vec x)$. For this, we take advantage of the following inequality:
\begin{align}
    ||\vec w(\vec x)|| = ||\mat {AB}\vec x|| \leq ||\mat{a}||_F\, ||\mat B\vec x|| \quad.
\end{align}

Here, $||\cdot||_F$ denotes the Frobenius norm and $||\cdot||$ the $L_2$ vector norm; this inequality reflects the fact that the Frobenius norm is \emph{compatible} with the $L_2$ vector norm. Note that using this approximation for the norm computation \emph{bounds the output bound} of the DAUs and is \emph{at least as tight} as the bound in eq.~\eqref{eq:bound}.
As a result, without the bias term $\vec b$, the output of the corresponding DAUs can be calculated as 
\begin{align}
    &\text{eDAU}(\vec{x})|| = ||\mat B\vec x||^{-1}\left(\mat{B}\vec x\right)^T \left(\mat A'^T\vec x\right)\quad
    \label{eq:eDAU}\\
    &\text{with} \quad \mat A' = ||\mat A||_F^{-1} \mat A 
\end{align}
We will henceforth refer to this non-linear output computation as \emph{weight bounding} (WB).
Note that under this formulation the $d$-dimensional weights $\vec w(\vec x)$ are never explicitly calculated and the output is instead obtained as a dot product in $\mathbb R^r$ between the vectors $\mat B\vec x$ and $\mat A'^T\vec x$. Further, for convolutional DAUs (see sec.~\ref{subsec:coda}), the matrix $\mat A'$ has to be computed only once for all positions. As we show in sec.~\ref{subsec:ablations}, this can result in significant gains in efficiency.
\subsection{DAUs for classification}
\label{subsec:classification}
{DAUs can be used directly for classification by
applying $k$ DAUs in parallel to obtain an output \mbox{$\hat{\vec y}(\vec x)=\left[\text{DAU}_1(\vec x), ..., \text{DAU}_k(\vec x)\right]$}. 
Note that this is a linear transformation $\hat{\vec y}(\vec x)$$=$$\mat W(\vec x) \vec x$, with each row in $\mat w$$\in$$\mathbb R^{k \times d}$ corresponding to the weight vector $\vec w_j^T$ of a specific DAU $j$.
Consider, for example,
a dataset \mbox{$\mathcal D = \{(\vec x_i, \vec y_i)|\, \vec x_i\in\mathbb R^d, \vec y_i\in\mathbb R^k\}$} of $k$ classes with `one-hot' encoded labels $\vec y_i$ for the inputs $\vec x_i$.
To optimise the DAUs as classifiers on $\mathcal D$,} we can apply a sigmoid non-linearity to each DAU output and optimise the loss function \mbox{$\mathcal L = \sum_i\text{BCE}(\sigma(\hat{\vec y}_i), \vec y_i)$}, where \text{BCE} denotes the binary cross-entropy and $\sigma$ applies the sigmoid function to each entry in $\hat{\vec y}_i$. Note that for a given sample, \text{BCE} either maximises (DAU for correct class) or minimises (DAU for incorrect classes) the output of each DAU. Hence, this classification loss will still maximise the (signed) cosine between the weight vectors $\vec w(\vec x_i)$ and $\vec x_i$. 

To illustrate this property, in Fig.~\ref{fig:alignment} (top) we show the weights $\vec w(\vec x_i)$ for several samples of the digit `3' after optimising the DAUs for classification on a noisy MNIST dataset; the first two are correctly classified, the last one is misclassified as a `5'. As can be seen, the weights align with the respective input (the weights for different samples are different). However,  different parts of the input are either positively or negatively correlated with a class, which is reflected in the weights: for example, the extended stroke on top of the `3' in the misclassified sample is assigned \emph{negative weight} and, since the background noise is \emph{uncorrelated} with the class labels, it is not represented in the weights. 

In a classification setting, the DAUs {thus} preferentially encode \emph{the most frequent discriminative patterns} in the linear transform $\mat{ab}\vec x + \vec b$ such that the dynamic weights $\vec w(\vec x)$ align well with these patterns.
{Additionally, since the output for class $j$ is a linear transformation of the input (\colornum{P1}), we can compute the contribution vector $\vec s_j$ containing the per-pixel contributions to this output by the element-wise product ($\odot$)
\begin{align}
\label{eq:contrib_1}
    \vec s_j(\vec x_i) = \vec w_j(\vec x_i)\odot\vec x_i\quad ,
\end{align} 
 see Figs.~\ref{fig:teaser} and
\ref{fig:alignment}. 
Such linear decompositions constitute the model-inherent `explanations' which we evaluate in sec.~\ref{sec:results}.}

\subsection{Convolutional Dynamic Alignment Networks}
\label{subsec:coda}
The modelling capacity of a single layer of DAUs is limited, similar to a single linear classifier. However, DAUs can be used as the basic building block for deep convolutional neural networks, which yields powerful classifiers. Importantly, in this section we show that such a Convolutional Dynamic Alignment Network (CoDA Net) inherits the properties (\colornum{P3}) of the DAUs by maintaining both the dynamic linearity (\colornum{P1}) as well as the alignment maximisation (\colornum{P2}). For a \emph{convolutional} dynamic alignment layer, each convolutional filter is modelled by a DAU, similar to dynamic local filtering layers~\cite{jia2016dynamic}. Note that the output of such a layer is also a dynamic linear transformation of the input to that layer, since a convolution is equivalent to a linear layer with certain constraints on the weights, cf.~\cite{convlin}. We include the implementation details in the supplement.
Finally, at the end of this section, we highlight an important difference between output maximisation and optimising for classification with the {BCE} loss. In this context we discuss the effect of \emph{temperature scaling} and present the loss function we optimise in our experiments.\\[-.75em]

\myparagraph{Dynamic linearity (\colornum{P1}).} In order to see that the linearity is maintained, we note that the successive application of multiple layers of DAUs also results in a dynamic linear mapping. Let $\mat W_l$ denote the linear transformation matrix produced by a layer of DAUs and let $\vec a_{l-1}$ be the input vector to that layer; as mentioned before, each row in the matrix $\mat w_l$ corresponds to the weight vector of a single DAU\footnote{
Note that this also holds for convolutional DAU layers. Specifically, each row in the matrix $\mat w_l$ corresponds to a single DAU applied to exactly one spatial location in the input and the input with spatial dimensions is vectorised to yield $\vec a_{l-1}$. For further details, we kindly refer the reader to~\cite{convlin} and the implementation details in the supplement of this work.}. As such, the output of this layer is given by 
\begin{align}
    \vec a_l = \mat W_l (\vec a_{l-1}) \vec a_{l-1}\quad .
\end{align}
In a network of DAUs, the successive linear transformations can thus be collapsed. In particular, \emph{for any pair of activation vectors} $\vec{a}_{l_1}$ and $\vec{a}_{l_2}$ with ${l_1}<{l_2}$, the vector $\vec{a}_{l_2}$ can 
    be expressed as a linear transformation of $\vec{a}_{l_1}$:
\begin{align}
\label{eq:collapse}
    \vec{a}_{l_2} &= \mat{W}_{{l_1}\rightarrow {l_2}} \left(\vec{a}_{l_1}\right)\vec{a}_{l_1} \quad 
        \\{with} \quad \mat{W}_{{l_1}\rightarrow {l_2}}\left(\vec{a}_{l_1}\right) &= \textstyle\prod_{k={l_1}+1}^{l_2} \mat{W}_k \left(\vec{a}_{k-1}\right)\quad \text{.}
\end{align}
For example, the matrix $\mat W_{0\rightarrow L}(\vec{a}_0 = \vec{x}) = \mat W(\vec{x})$ models the linear transformation from the input to the output space, see Fig.~\ref{fig:teaser}.
Since this linearity holds between any two layers, the $j$-th entry of any activation vector $\vec a_l$ in the network can be decomposed into input contributions via:
    \begin{align}
    \label{eq:contrib}
        \vec{s}_{j}^l(\vec x_i) = \left[\mat W_{0\rightarrow l} (\vec{x}_i)\right]_j^T \odot \vec x_i\quad \text{,}
    \end{align}
    with $[\mat W]_j$ the $j$-th row in the matrix.

\myparagraph[-2]{Alignment maximisation (\colornum{P2}).}
Note that the output of a CoDA Net is bounded independent of the network parameters: since each DAU operation can---independent of its parameters---at most reproduce the norm of its input (eq.~\eqref{eq:bound}), the linear concatenation of these operations necessarily also has an upper bound which does not depend on the parameters.
Therefore, in order to achieve maximal outputs on average (e.g., the class logit over the subset of images of that class), all DAUs in the network need to produce weights $\vec w (\vec a_l)$ that align well with the class features. In other words, the weights will align with discriminative patterns in the input.
For example, in Fig.~\ref{fig:alignment} (bottom), we visualise the `global matrices' $\mat W_{0\rightarrow L}$ and the corresponding contributions (eq.~\eqref{eq:contrib}) for a $L=5$ layer CoDA Net. As before, the weights align with discriminative patterns in the input and do not encode the uninformative noise.

\myparagraph[0]{Temperature scaling and loss function.} %
\begin{figure}[t]
    \centering
    \includegraphics[width=.45\textwidth]{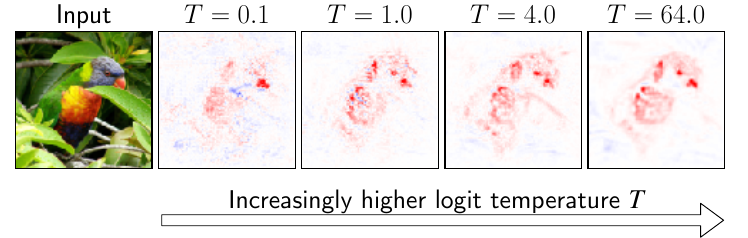}
    \caption{\small By lowering the upper bound (cf.~eq.~\eqref{eq:bound}), the correlation maximisation in the DAUs can be emphasised.
    We show contribution maps for a model trained with different temperatures. 
    }
    \label{fig:scaling}
\end{figure}
So far we have assumed that minimising the {BCE} loss for a given sample is equivalent to applying a maximisation or minimisation loss to the individual outputs of a CoDA Net. While this is in principle correct, {BCE} introduces an additional, non-negligible effect: \emph{saturation}. Specifically, it is possible for a CoDA Net to achieve a low {BCE} loss without the need to produce well-aligned weight vectors. As soon as the classification accuracy is high and the outputs of the networks are large, the gradient---and therefore the \emph{alignment pressure}---will vanish. This effect can, however, easily be mitigated:
 as discussed in the previous paragraph, the output of a CoDA Net is upper-bounded \textit{independent of the network parameters}, since each individual DAU in the network is upper-bounded. 
By scaling the network output with a temperature parameter $T$ such that 
    $\hat{\vec y} (\vec x) = T^{-1} \mat W_{0\rightarrow L}(\vec x)\,\vec x$, 
we can explicitly decrease this upper bound and thereby increase the \emph{alignment pressure} in the DAUs by avoiding the early saturation due to {BCE}.
In particular, the lower the upper bound is, the stronger the induced DAU output maximisation should be, since the network needs to accumulate more signal to obtain large class logits (and thus a negligible gradient). This is indeed what we observe both qualitatively, cf.~Fig.~\ref{fig:scaling}, and quantitatively, cf.~Fig.~\ref{fig:localisation} (right column).
The overall loss for an input $\vec x_i$ and the target vector $\vec y_i$ is thus computed as 
    \begin{align}
        \label{eq:loss}
        \mathcal{L}(\vec x_i, \vec y_i) &= 
        \text{BCE}(\sigma(T^{-1} \mat W_{0\rightarrow L}(\vec x_i)\,\vec{x}_i + {\vec{b}}_0)\,,\, \vec{y}_i)\quad \text{.}
    \end{align}
    Here,     $\sigma$ applies the sigmoid activation to each vector entry and
    ${\vec{b}}_0$ is a fixed bias term.
        As an alternative to the temperature scaling, the explicit representation of the network's computation as a linear mapping allows to directly regularise what properties these linear mappings should fulfill. For example, we show in the supplement that by regularising the absolute values of the matrix $\mat W_{0\rightarrow L}$, we can induce sparsity in the signal alignments, which also leads to sharper heatmaps.
\section{Experimental setup}
\label{sec:experiments}

\subsection{Datasets}
\label{subsec:datasets}
We evaluate and compare the accuracies of the CoDA Nets to other work on the CIFAR-10~\cite{krizhevsky2009cifar10} and the TinyImagenet~\cite{tinyimagenet} datasets. We use the same datasets for the quantitative evaluations of the model-inherent contribution maps. Additionally, we qualitatively show high-resolution examples from a CoDA Net trained on the first 100 classes of the Imagenet~\cite{imagenet} dataset. Lastly, we evaluate hybrid models (see sec.~\ref{subsec:interpolation}) on CIFAR10 and the full Imagenet dataset, both in terms of interpretability as well as classification accuracy.
\subsection{Models}
Our results (secs.~\ref{subsec:accuracy}--\ref{subsec:ablations})
are based on models of various sizes denoted by (S/L/XL)-CoDA on CIFAR-10 (S), Imagenet-100 (L), and TinyImagenet (XL); these models have 7-8M\footnote{
The models with SQ and L2 non-linearity have 7.8M parameters and the models with WB have 7.1M (without embedding) and 7.2M (with embedding) parameters.
} (S), 48M (L), and 62M (XL) parameters respectively; {see the supplement for details on the model architectures and an evaluation of the impact of model size on accuracy}. 
For the hybrid networks (see secs.~\ref{subsec:interpolation} and~\ref{subsec:interpolation_results}), we use a ResNet-56 (ResNet-50) as a base model on CIFAR-10 (Imagenet) and train CoDA Nets on feature maps extracted at different depths of the base models; see the supplement for details.

\subsection{Input encoding}
\label{subsec:encoding}
In sec.~\ref{subsec:align_units}, we discussed that the norm-weighted cosine similarity between the dynamic weights and the layer inputs is optimised and the output of a DAU is at most the norm of its input. 
When using pixels as the input to the CoDA Nets, this favours pixels with large RGB values, since these have a larger norm and can thus produce larger outputs in the maximisation task.
In our experiments, we explore two approaches to mitigate this bias: in the first, we add the negative image as three additional color channels and thus encode each pixel in the input 
as \mbox{[$r$, $g$, $b$, $1-r$, $1-g$, $1-b$]}, with $r, g, b\in [0, 1]$.

Secondly, we show that it is also possible to train CoDA Nets on end-to-end optimised patch-embeddings and obtain similar performance in terms of interpretability and classification accuracy. Instead of computing the \emph{per-pixel} contributions to  assess the importance of spatial locations (cf.~eq.~\eqref{eq:contrib}), in our experiments we thus  decompose the output with respect to the contributions from the corresponding (learnt) embeddings via
\begin{align}
    \label{eq:contrib_emb}
        \vec{s}_{j}^L(\vec x_i) = \left[\mat W_{0\rightarrow L} (E(\vec{x}_i))\right]_j^T \odot E(\vec x_i)\quad \text{,}
    \end{align}
    with $E(\cdot)$ denoting the applied embedding function and $L$ the number of CoDA layers in the network.
\subsection{Interpolating between networks}
\label{subsec:interpolation}
Training CoDA Nets on learnt patch-embeddings (see sec.~\ref{subsec:encoding}) naturally raises the question of how complex the embedding function should be and how large its receptive field. In particular, are \emph{pixel-wise} importance values more useful than importance values for embeddings of patches of size $3\times 3$? How about $7\times 7$ or $64\times 64$? Of course, there is no single answer to this question and the `optimal' complexity of the embedding model depends on the dataset, the task, and, ultimately, on the preferences of the end-user of such a model: for example, if a more complex embedding allows for more performant classifiers, one might wish to trade off model interpretability against model accuracy.
In order to better understand such trade-offs, we propose to `interpolate' between a conventional CNN and the CoDA Nets and investigate how this affects both model interpretability and model performance. Specifically, starting from a pre-trained CNN, we successively replace an increasing number of the later layers of the base model by CoDA layers. As such, the model output can be decomposed into \emph{contributions coming from spatially arranged embeddings} computed by the truncated CNN model, which can give insights into how the embeddings are used to produce the classification results.

\subsection{Additional details}
\label{subsec:shared_B}

In our experiments, we observed that rescaling the weight vectors of the DAUs explicitly according to eq.~\eqref{eq:nonlin} resulted in long training times and high memory usage. To mitigate this, we opted to share the matrix $\mat B$ between all DAUs in a given layer when using the L2 or SQ non-linearity. This increases efficiency by having the DAUs share a common $r$-dimensional subspace and still fixes the maximal rank of each DAU to the chosen value of $r$. In contrast, networks with the WB non-linearity (see eDAUs in eq.~\eqref{eq:eDAU}) are specifically designed to lower the computational costs of the DAUs and are easier to train. Therefore, for CoDA Nets built with eDAUs, we do not share the matrices $\mat b$ between the eDAUs. As the inputs are thus not restricted to a common low-dimensional subspace, we expect this to increase the modelling capacity of the CoDA Nets.

\section{Experiments}
\label{sec:results}
\begin{figure*}[t]
\centering
\fcolorbox[rgb]{0.43921569, 0.50196078, 0.56470588}{0.9176470588235294, 0.9176470588235294, 0.9490196078431372}{
\begin{subfigure}[c]{.925\textwidth}
\includegraphics[width=1\textwidth]{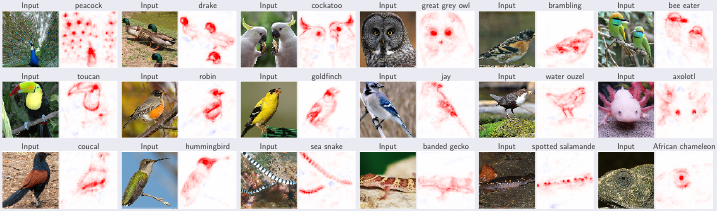}
\end{subfigure}
}
            \caption{\small Model-inherent contribution maps for the most confident predictions for 18 different classes, sorted by confidence (high to low). We show positive (negative) contributions (eq.~\eqref{eq:contrib}) per spatial location for the ground truth class logit in red (blue).}
            \label{fig:quality}
\vspace{-1em}
\end{figure*}
In sec.~\ref{subsec:accuracy} we assess the classification performance of the CoDA Nets. 
Further, in sec.~\ref{subsec:intp_results} we evaluate the model-inherent contribution maps derived from $\mat w_{0\rightarrow L}$ (cf.~eq.~\eqref{eq:contrib_emb}) of a CoDA Net and compare them both \emph{qualitatively} (cf.~Fig.~\ref{fig:quality}) as well as \emph{quantitatively} (cf.~Fig.~\ref{fig:localisation}) to other attribution methods. 
Additionally, in sec.~\ref{subsec:ablations}, we discuss the impact of the different rescaling methods (cf.~eqs.~\eqref{eq:nonlin} and~\eqref{eq:eDAU}) 
on model interpretability and evaluation speed.
Lastly, in sec.~\ref{subsec:interpolation_results} we investigate the hybrid models discussed in sec.~\ref{subsec:interpolation}. In particular, we analyse how the depth of the embedding function $E(\vec x)$ affects the model interpretability at different depths of the resulting hybrid network architecture.

\subsection{Model performance}
\label{subsec:accuracy}
\begin{table}
    \centering
    {\setlength{\tabcolsep}{0.25em}\setlength\extrarowheight{-1pt}
    \begin{tabular}{>{\centering\arraybackslash} >{\centering\arraybackslash}p{2.85cm} | >{\centering\arraybackslash}p{.75cm} >{\centering\arraybackslash}p{.1cm}  >{\centering\arraybackslash}p{3cm} | >{\centering\arraybackslash}p{.75cm} }
    \footnotesize \textbf{Model} & \footnotesize \textbf{C10}
    && \footnotesize \textbf{Model} & \footnotesize \textbf{T-IM}\\[.2em]
    \cline{1-2}
    \cline{4-5}
    & \footnotesize&& & \footnotesize\\[-.8em]
    \footnotesize SENNs~\cite{melis2018towards} & \footnotesize 78.5\% &
    & \footnotesize ResNet-34~\cite{resnet-tiny} & \footnotesize 52.0\%
    \\
    \footnotesize DE-CapsNet~\cite{jia2020capsnet} & \footnotesize 93.0\%&
    & \footnotesize VGG~16~\cite{vgg-tiny} & \footnotesize 52.2\%
    \\
    \footnotesize VGG-19~\cite{li2020few} & \footnotesize 93.4\%&
    & \footnotesize VGG~16  + aug \cite{vgg-tiny} & \footnotesize 56.4\%
    \\
    \footnotesize ResNet-110~\cite{he2016deep} & \footnotesize 93.6\%&
    & \footnotesize IRRCNN~\cite{irrcnn-alom} & \footnotesize 52.2\%
    \\
    \footnotesize DenseNet \cite{huang2017densely} & \footnotesize 94.8\%&
    & \footnotesize ResNet-110~\cite{rn110-tiny} & \footnotesize 56.6\%
    \\
    \footnotesize WRN-28-2~\cite{zagoruyko2016wide} & \footnotesize 94.9\%&
    & \footnotesize WRN-40-20~\cite{hendrycks-tiny} & \footnotesize 63.8\%
    \\[.2em]
    \cline{1-2}\cline{4-5}
    &&&&\\[-.75em]
    \footnotesize S-CoDA-SQ& \footnotesize 93.2\%&
    & \footnotesize XL-CoDA-SQ & \footnotesize 54.4\%
    \\
    \footnotesize S-CoDA-L2& \footnotesize 93.0\%&
    & \footnotesize XL-CoDA-SQ + aug & \footnotesize 58.4\%
    \\
    \footnotesize S-eCoDA-WB & \footnotesize 94.0\%&
     &&
     \\
    \footnotesize S-eCoDA-WB + $E(\vec x)$& \footnotesize 94.1\%&
    & &
    \end{tabular}}
    \caption{\small CIFAR-10 (\textbf{C10}) and TinyImagenet (\textbf{T-IM}) 
    classification accuracies. Results taken from specified references. The prefix of the CoDAs indicates model size, the suffix the non-linearity used (cf.~eqs.~\eqref{eq:nonlin} and \eqref{eq:eDAU}). Further, $E(\vec x)$ denotes that a learnt embedding was used as input to the model (see sec.~\ref{subsec:encoding}).}
    \label{tbl:result_table}
    \vspace{-1.75em}
\end{table}
\myparagraph{Classification performance.} In Table \ref{tbl:result_table} we compare the performances of our CoDA Nets to several other published results. Note that the referenced numbers are meant to be used as a gauge for assessing the CoDA Net performance and do not exhaustively represent the state of the art. In particular, we would like to highlight that the CoDA Net performance is on par to that of models from the VGG~\cite{vgg} and ResNet~\cite{he2016deep} model families on both datasets. 
Additionally, we list the reported results of the SENNs~\cite{melis2018towards} and the DE-CapsNet~\cite{jia2020capsnet} architectures for CIFAR-10. Similar to our CoDA Nets, the SENNs were designed to improve network interpretability and are also based on the idea of explicitly modelling the output as a dynamic linear transformation of the input. On the other hand, the CoDA Nets share similarities to capsule networks, which we discuss in the supplement; to the best of our knowledge, the \mbox{DE-CapsNet} currently achieves the state of the art in the field of capsule networks on CIFAR-10. 
Overall, we observed that the CoDA Nets deliver competitive performances that are fairly robust to the non-linearity (see eqs.~\eqref{eq:nonlin} and~\eqref{eq:eDAU}) and the temperature ($T$); for an ablation study on the latter, see the supplement. 
Finally, while all models achieve good classification results, we note that the WB-based CoDA Nets perform slightly better than CoDA Nets with SQ or L2 non-linearity despite having a comparable amount of parameters. As discussed in sec.~\ref{subsec:shared_B}, we attribute this to the fact that for those models we do not share the matrix $\mat b$ within the CoDA layers, which increases their modelling capacity.

\subsection{Interpretability of CoDA Nets}
\label{subsec:intp_results}
\vspace{-.2em}
\begin{figure*}
%
    \centering
    \begin{subfigure}[c]{0.325\textwidth}
    \centering
    \includegraphics[width=\textwidth, trim=0 1em 0 1em, clip]{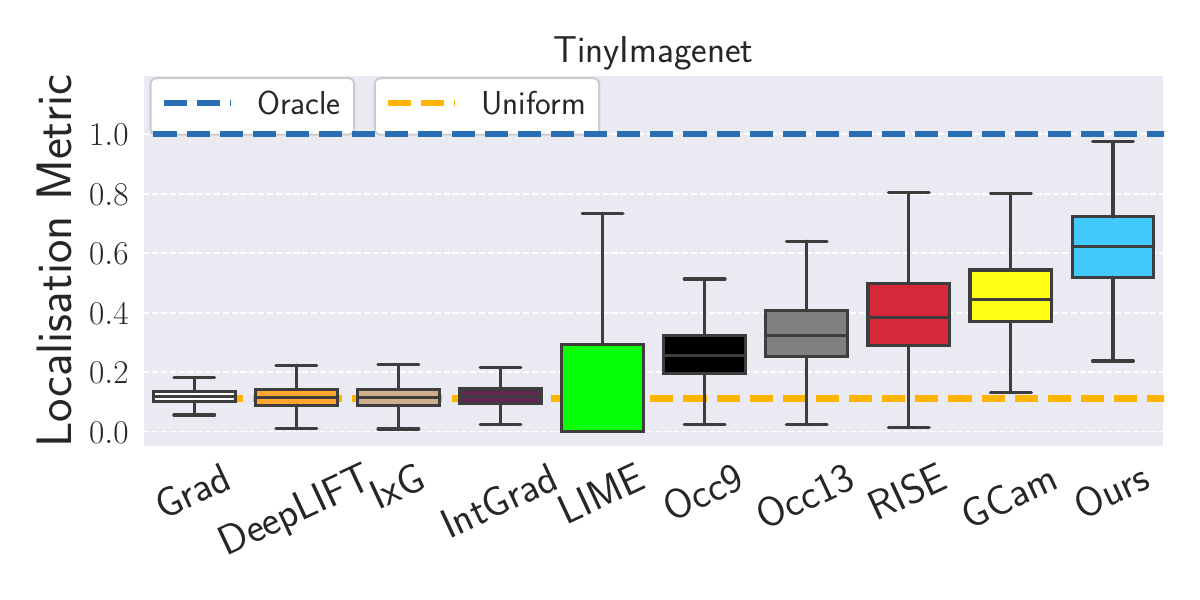}
    \end{subfigure}
    \begin{subfigure}[c]{0.325\textwidth}
    \centering
    \includegraphics[width=\textwidth, trim=0 1em 0 1em, clip]{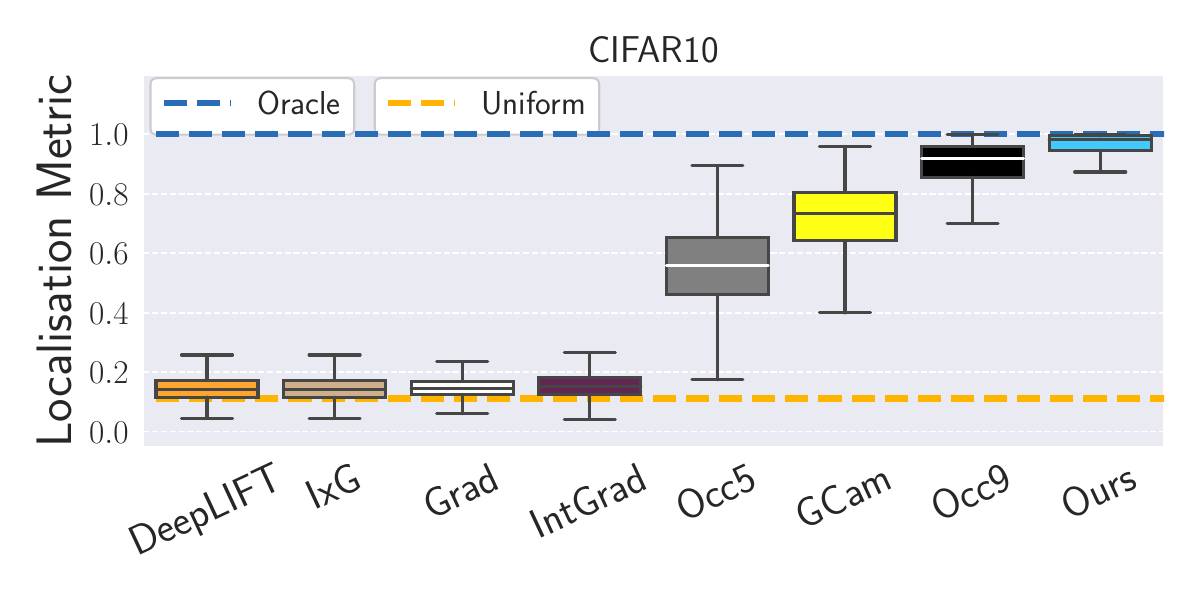}
    \end{subfigure}
    \begin{subfigure}[c]{0.325\textwidth}
    \centering
    \includegraphics[width=\textwidth, trim=0 1em 0 1em, clip]{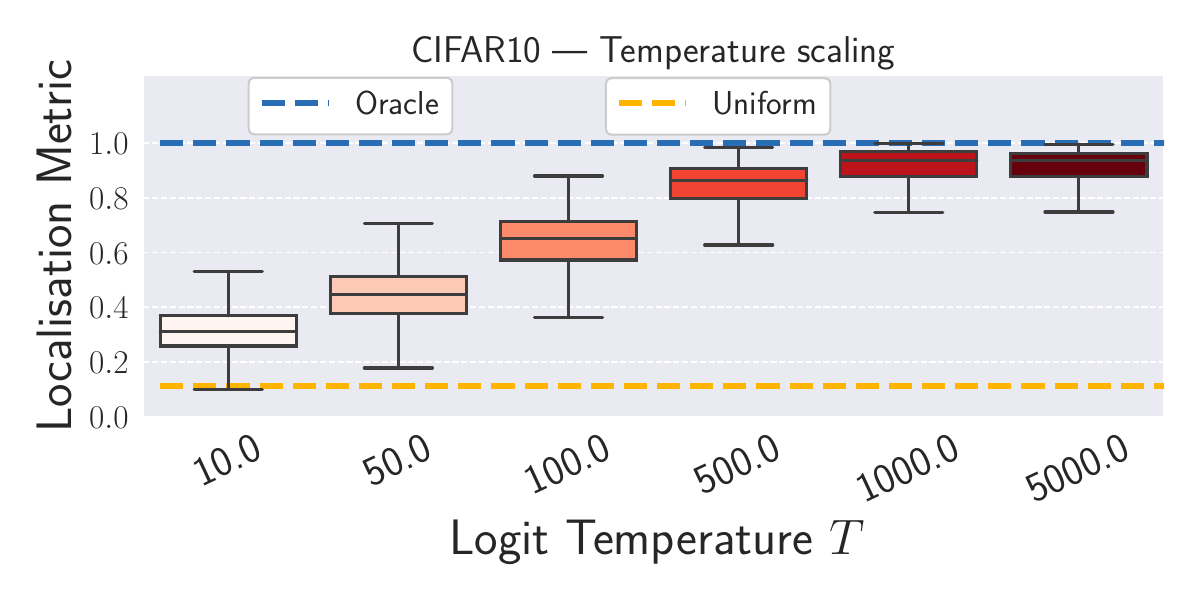}
    \end{subfigure}
    \begin{subfigure}[c]{0.325\textwidth}
    \centering
    \includegraphics[width=\textwidth, trim=0 1em 0 1em, clip]{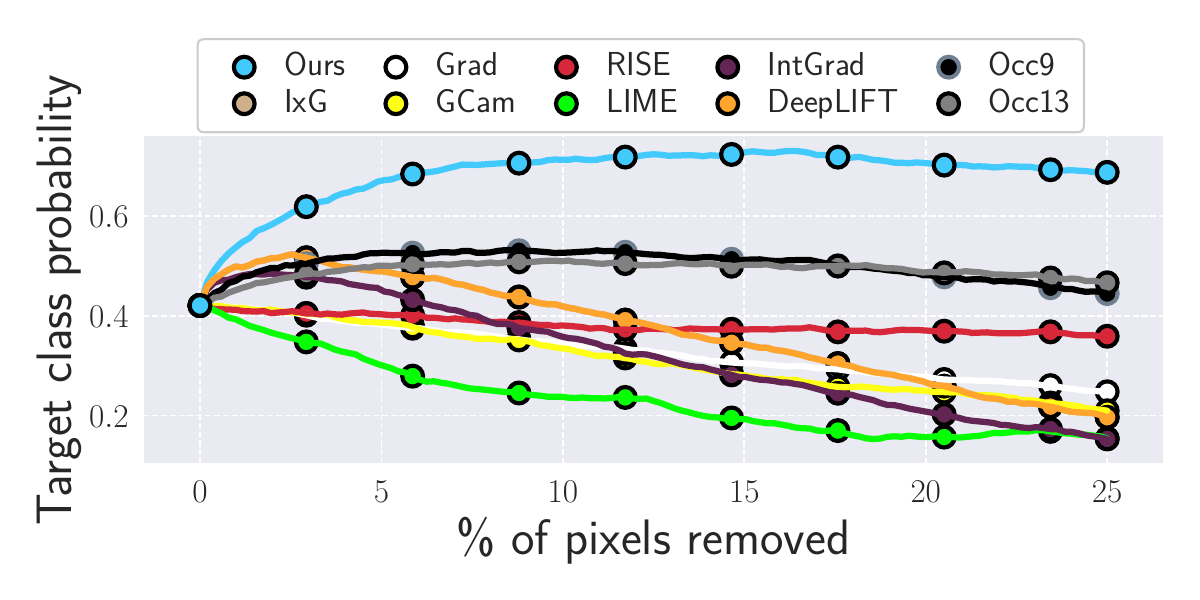}
    \end{subfigure}
    \begin{subfigure}[c]{0.325\textwidth}
    \centering
    \includegraphics[width=\textwidth, trim=0 1em 0 1em, clip]{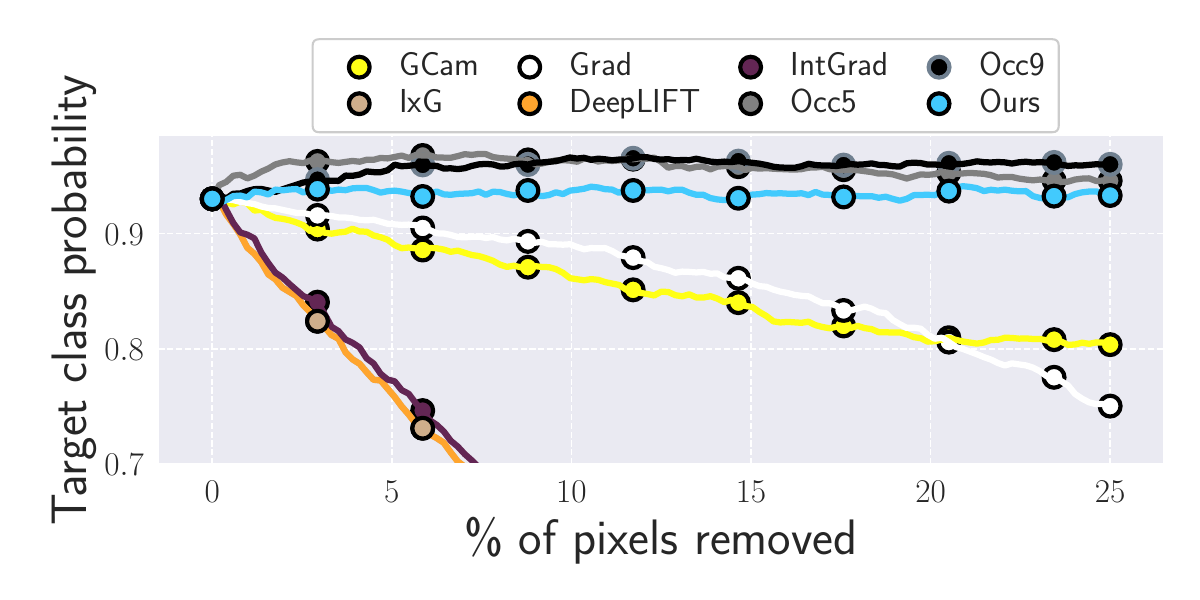}
    \end{subfigure}
    \begin{subfigure}[c]{0.325\textwidth}
    \centering
    \includegraphics[width=\textwidth, trim=0 1em 0 1em, clip]{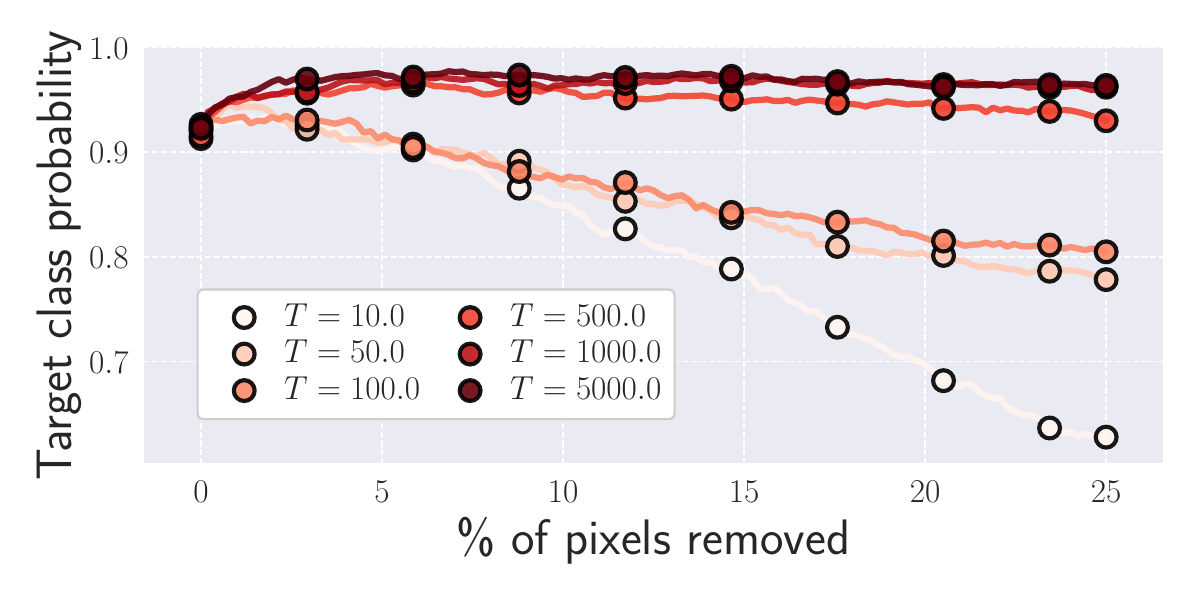}
    \end{subfigure}
    \caption{\small 
    \textbf{Top row:} Results for the localisation metric, see eq.~\eqref{eq:loc_metric}.
    \textbf{Bottom row:} Pixel removal metric. In particular, we plot the mean target class probability  after removing the $x\%$ of the \emph{least important} pixels.
    We show the results of a CoDA-Net-SQ trained on TinyImagenet \textbf{(left column)}, as well as of a CoDA-Net-WB trained on CIFAR-10 \textbf{(center column)}. We observed the results for different rescaling methods (SQ/L2/WB) to be very similar and therefore just show one per dataset; more results can be found in the supplement. 
    Additionally, we show the effect of the temperature parameter on the interpretability of a CoDA-Net with SQ rescaling \textbf{(right column)}:
    as expected, a higher temperature leads to higher interpretability (sec.~\ref{subsec:coda}). 
    }
    \label{fig:localisation}
\end{figure*}
In the following, we evaluate the model-inherent contribution maps and compare them to other commonly used me\-thods for importance attribution.
The evaluations are based on the XL-CoDA-SQ (T=6400) for TinyImagenet and the S-eCoDA-WB (T=$1e6$) for CIFAR-10, see~Table~\ref{tbl:result_table} for the respective accuracies.
The results are very similar for all three non-linearities (cf.~sec.~\ref{subsec:ablations}; more results are included in the supplement) and we therefore just show one per dataset as an example.
Further, we evaluate the effect of training the S-CoDA-SQ architecture with different temperatures $T$; as discussed in sec.~\ref{subsec:coda}, we expect the interpretability to \emph{increase} along with $T$, since for larger $T$ a stronger alignment is required in order for the models to obtain large class logits. 
Lastly, in sec.~\ref{subsec:ablations} we compare how the different non-linearities (L2, SQ, WB) affect the model interpretability.
Before turning to the results, however, in the following we will first present the attribution methods used for comparison and discuss the evaluation metrics employed for quantifying their interpretability.

\myparagraph{Attribution methods.} We compare the model-inherent contribution maps (cf.~eq.~\eqref{eq:contrib}) to other common approaches for importance attribution. 
In particular, we evaluate against several perturbation based methods such as RISE~\cite{petsiuk2018rise}, LIME~\cite{lime}, and several occlusion attributions~\cite{zeiler2014visualizing} (Occ-K, with K the size of the occlusion patch). Additionally, we evaluate against common gradient-based methods. These include the gradient of the class logits with respect to the input image~\cite{baehrens2010explain} (Grad), `Input$\times$Gradient' (IxG, cf.~\cite{adebayo2018sanity}), GradCam~\cite{selvaraju2017grad} (GCam), Integrated Gradients~\cite{sundararajan2017axiomatic} (IntG), and DeepLIFT~\cite{shrikumar2017deeplift}. As a baseline, we also evaluated these methods on a pre-trained ResNet-56~\cite{he2016deep} on CIFAR-10, for which we show the results in the supplement.

\begin{figure}
    \centering
    \fcolorbox[rgb]{0.43921569, 0.50196078, 0.56470588}{0.9176470588235294, 0.9176470588235294, 0.9490196078431372}
    {
    \centering
    \includegraphics[width=.45\textwidth, trim=2em 1em 1em 0em]{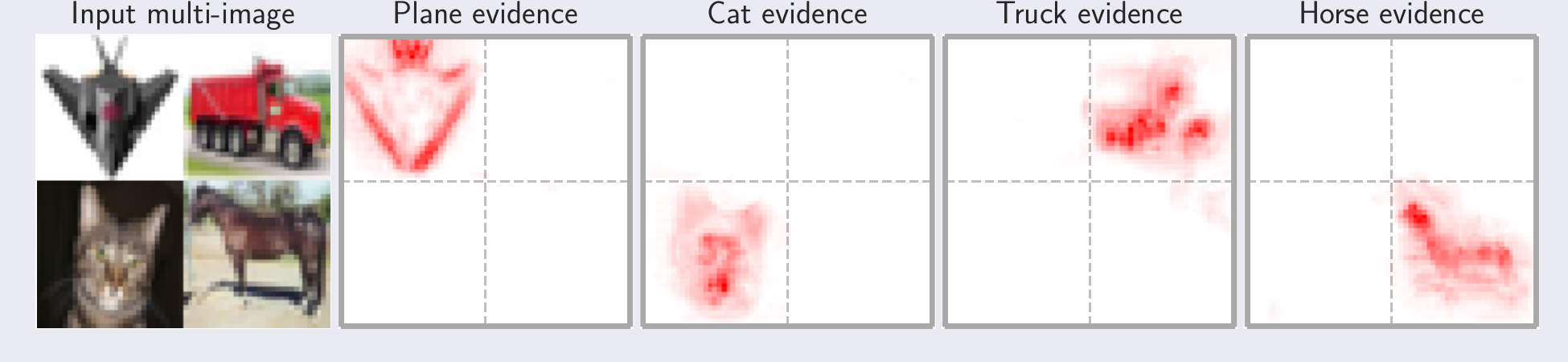}
    }
    \caption{\small A multi-image on the CIFAR-10 dataset. The CoDA-Net contribution maps highlight the individual class-images well.}
    \label{fig:multi_image}
\end{figure}

\myparagraph{Evaluation metrics.}
Our quantitative evaluation of the attribution maps is based on the following two methods:
we
    \mbox{\colornum{(1)} evaluate} a localisation metric by adapting the pointing game~\cite{zhang2018top} to the CIFAR-10 and TinyImagenet datasets, and 
    \mbox{\colornum{(2)} analyse} the model behaviour under the pixel removal strategy employed in~\cite{srinivas2019full}.
For~\colornum{(1)}, 
we evaluate the attribution methods on a grid of $ n\times n$ with $n=3$ images sampled from the corresponding datasets; in every grid of images, each class may occur at most once. 
For a visualisation with $n=2$, see
Fig.~\ref{fig:multi_image}.
\begin{figure}
\centering
\fcolorbox[rgb]{0.43921569, 0.50196078, 0.56470588}{0.9176470588235294, 0.9176470588235294, 0.9490196078431372}{
    \begin{subfigure}[c]{.45\textwidth}
    \centering
    \includegraphics[width=\textwidth]{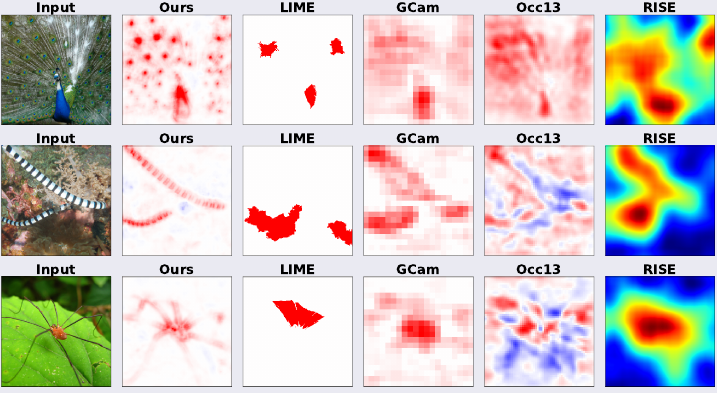}
            \end{subfigure}}
            \caption{\small Comparison to the strongest post-hoc methods. While the regions of importance roughly coincide, the inherent contribution maps of the CoDA-Nets offer the most detail. Note that to improve the RISE visualisation, we chose its default colormap~\cite{petsiuk2018rise}; the most (least) important values are still shown in red (blue).}
            \label{fig:comparison}
\end{figure}
For each occurring class, we can measure how much positive importance an attribution method assigns to the respective class image.
Let $\mathcal{I}_c$ be the image for class $c$, then the score $s_c$ for this class is calculated as 
\begin{align}
\label{eq:loc_metric}
            \textstyle s_c = \frac{1}{Z}\sum_{p_c \in \mathcal{I}_c} p_c \quad \text{with}\quad 
            Z = {\sum_k\sum_{p_c \in \mathcal{I}_k} p_c}\quad ,
        \end{align}
        with $p_c$ the positive attribution for class $c$ assigned to the spatial location $p$.
        This metric has the same clear oracle score  $s_c=1$ for all attribution methods (all positive attributions located in the correct grid image)
        and a clear score for completely random attributions $s_c=1/n^2$ (the positive attributions are uniformly distributed over the different grid images).
        Since this metric depends on the classification accuracy of the models, we sample the first $500$ (CIFAR-10) or $250$ (TinyImagenet) images according to their class score for the ground-truth class\footnote{
        We can only expect an attribution to specifically highlight a class image if this image can be correctly classified on its own. If all grid images have similarly low attributions, the localisation score will be random.
        }; note that since all attributions are evaluated for the same model on the same set of images, this does not favour any particular attribution method.\\
For~\colornum{(2)}, we show how the model's class score behaves under the removal of an increasing amount of \emph{least important} pixels,
    where the importance is obtained via the respective attribution method. 
Since the first pixels to be removed are typically assigned negative or relatively little importance, we expect the model to initially increase its confidence (removing pixels with \emph{negative} impact) or maintain a similar level of confidence (removing pixels with \emph{low} impact) if the evaluated attribution method produces an accurate ranking of the pixel importance values. 
Conversely, if we were to remove the \emph{most important} pixels first, we would expect the model confidence to quickly decrease. However, as noted by \cite{srinivas2019full}, removing the most important pixels first introduces artifacts in the most important regions of the image and is therefore potentially more unstable than removing the least important pixels first.
Nevertheless, the model-inherent contribution maps perform well in this setting, too, as we show in the supplement.
{Lastly, in the supplement we qualitatively show that they pass the `sanity check' of \cite{adebayo2018sanity}.}

\myparagraph[0]{Quantitative results.}
In Fig.~\ref{fig:localisation}, we compare the 
contribution maps of the CoDA Nets to other attributions under the evaluation metrics discussed above.
It can be seen that the CoDA Nets~\colornum{(1)}
    perform well under the localisation metric given by eq.~\eqref{eq:loc_metric} and outperform all 
    the other attribution methods evaluated on the same model, both for TinyImagenet (top row, left) and CIFAR-10 (top row, center); note that we excluded RISE and LIME on CIFAR-10, since the default parameters do not seem to transfer well to this low-resolution dataset. 
Moreover, \colornum{(2)} 
    the CoDA Nets perform well in the pixel-removal setting: 
    the \emph{least salient} locations according to the model-inherent contributions indeed seem to be among the least relevant for the given class score on both datasets, see Fig.~\ref{fig:localisation} (bottom row, left and center); note that the Occ-K explanations directly estimate the impact of occluding pixels and are thus expected to perform well under this metric.
Further, in Fig.~\ref{fig:localisation} (right column),  we show the effect of temperature scaling on the interpretability of CoDA Nets with SQ rescaling trained on CIFAR-10. The results indicate that the alignment maximisation is indeed crucial for interpretability and constitutes an important difference of the CoDA Nets to other dynamic linear networks such as piece-wise linear networks (ReLU-based networks). In particular, by structurally requiring a strong alignment for confident classifications, the interpretability of the CoDA Nets forms part of the optimisation objective.
Increasing the temperature increases the alignment and thereby the interpretability of the CoDA Nets. {While we observe a downward trend in classification accuracy when increasing $T$, the best model at $T=10$ only slightly improved the accuracy compared to $T=1000$ ($93.2\%\rightarrow 93.6\%$); for more details, see supplement.}

In summary, the results show that by combining dynamic linearity with a structural bias towards an alignment with discriminative patterns, we obtain models which inherently provide an interpretable linear decomposition of their predictions. 
Further, given that we better understand the relationship between the intermediate computations and the optimisation of the final output in the CoDA Nets, we can emphasise model interpretability in a principled way by increasing the `alignment pressure' via \emph{temperature scaling}.
    
\myparagraph{Qualitative results.} In Fig.~\ref{fig:quality}, we visualise spatial contribution maps of an L-CoDA-SQ model (trained on Imagenet-100) for some of its most confident predictions. Note that these contribution maps are linear decompositions of the output and the sum over these maps yields the respective class logit. In Fig.~\ref{fig:comparison}, we additionally present a visual comparison to the best-performing post-hoc attribution methods; note that RISE cannot be displayed well under the same color coding and we thus use its default visualisation. We observe that the different methods are not inconsistent with each other and roughly highlight similar regions. However, the inherent contribution maps are of much higher detail and compared to the perturbation-based methods do not require multiple model evaluations. Much more importantly, however, all the other methods are attempts at approximating the model behaviour \emph{post-hoc}, while the CoDA Net contribution maps in Fig.~\ref{fig:quality} are derived from the model-inherent linear mapping that is used to compute the model output.

\subsection{Interpretability and efficiency of L2, SQ, and WB}
\label{subsec:ablations}
In the following, we present results regarding the impact of the different normalisation methods (L2, SQ, WB) on model interpretability and model efficiency.
\begin{figure*}
\centering
    \begin{subfigure}[c]{.475\textwidth}
    \centering
    \includegraphics[width=1\textwidth]{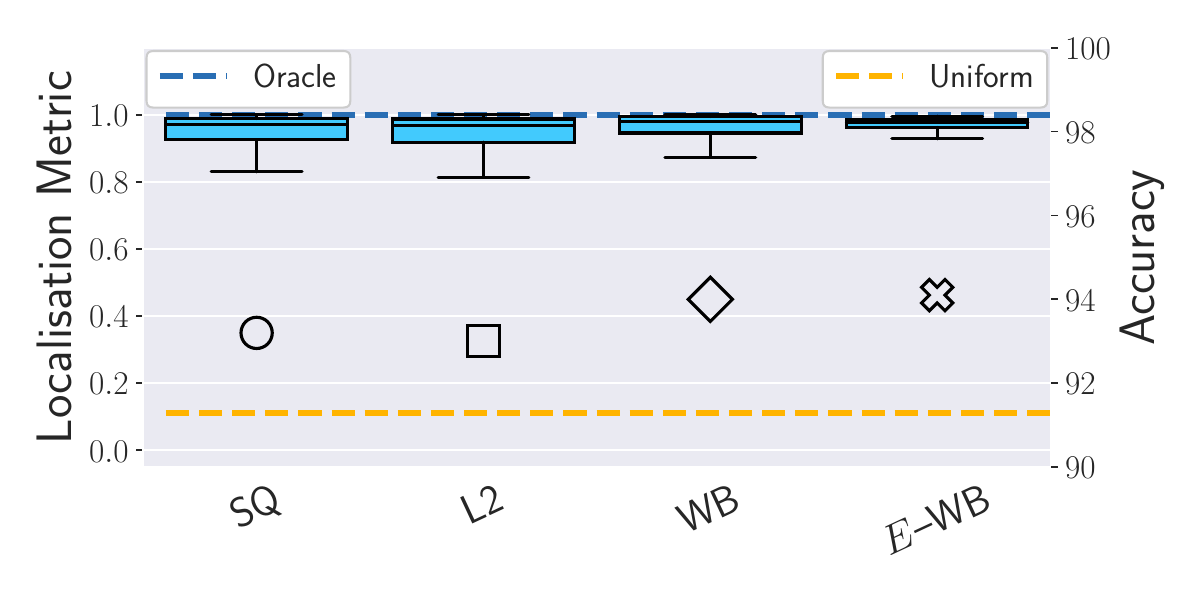}
    \end{subfigure}
    \begin{subfigure}[c]{.475\textwidth}
    \centering
    \includegraphics[width=1\textwidth]{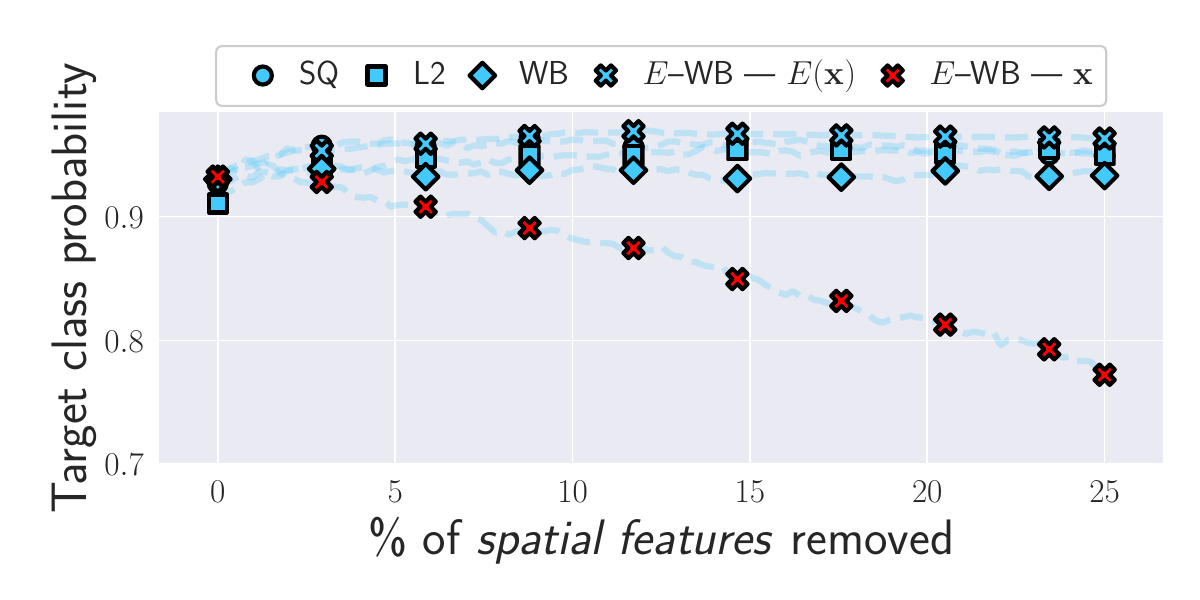}
    \end{subfigure}
    \caption{\small Localisation (\textbf{left}) and perturbation (\textbf{right}) metric results on CIFAR10 evaluated for CoDA Networks with different non-linearities (cf.~eqs.~\eqref{eq:nonlin} and \eqref{eq:eDAU}) 
    as well as trained with a learnt patch-embedding $E(\vec x)$, here denoted by $E$--WB. 
    For the model with embedding function $E(\vec x)$, we evaluate the pixel perturbation metric (bottom) directly on the pixels (red crosses) as well as on the learnt patch-embeddings (blue crosses).
    We further added the models' accuracies (see Table \ref{tbl:result_table}) in the plot to the left for comparison (circle / square/ cross markers). }
    \label{fig:nonlin_comparison}
\end{figure*}

\myparagraph{Model interpretability.} In Fig.~\ref{fig:nonlin_comparison}, we show the results of the interpretability metrics for models with different rescaling functions (L2/SQ/WB, see eqs.~\eqref{eq:nonlin} and~\eqref{eq:eDAU}) as well as for a model trained with a learnt patch-embedding $E(\vec x)$. 
As an embedding function, we simply apply a 3x3 convolution with 32 filters, followed by a batch normalisation layer~\cite{ioffe2015batch}. For comparison to post-hoc methods evaluated on a CoDA Net, we kindly refer the reader to the center column of Fig.~\ref{fig:localisation}. As can be seen, it is possible to obtain highly interpretable models under all
four settings: \colornum{(1)} the linear contributions allow to localise the class-images well (localisation metric, left) and \colornum{(2)} the models are insensitive to \emph{\bf input features}\footnote{For the SQ, L2, and WB model the features are pixels under the static encoding function described in \ref{subsec:encoding}. For the $E$-WB model, the input features to the CoDA Net are \emph{learnt} patch embeddings.} that are not contributing to the output as per the linear transformation matrix $\mat W_{0\rightarrow L}$. Note that for the model with a learnt patch-embedding, denoted by $E$--WB, we show two results for the perturbation metric. First, we 'zero out' the \emph{embeddings} at each location ordered by their assigned importance (blue crosses). As the embeddings are the input features to the CoDA Net, the model confidence shows the expected behaviour of being insensitive to unimportant inputs.
In contrast, the assigned importance values do not translate to the center pixels of the embeddings: when zeroing out the \emph{center pixels} according to the contributions of the patch-embeddings, the model confidence drops more quickly (see red crosses). This distinction is important to keep in mind when evaluating CoDA Nets on input embeddings,  
since it is easy to wrongly interpret such contribution maps. If the input to the CoDA Net is an embedding of an image patch, it depends on the embedding function how the contributions are to be distributed to the image pixels.
Lastly, note that different from the center column in Fig.~\ref{fig:localisation}, the metrics are evaluated for \emph{four different models} and are thus not comparisons between different explanation methods, but rather between different models under the same explanation. As such, the  differences in the localisation metric, for example, do not show that the linear decompositions are generally better suited to explain WB-based models as compared to SQ- or L2-based models; the differences might instead reflect the fact that the models learnt more robust and class-specific representations, which yield both better results in the localisation task as well as higher classification accuracy. 

\myparagraph{Model efficiency.}
\begin{figure*}
\centering
    \begin{subfigure}[c]{.45\textwidth}
    \centering
    \includegraphics[width=1\textwidth]{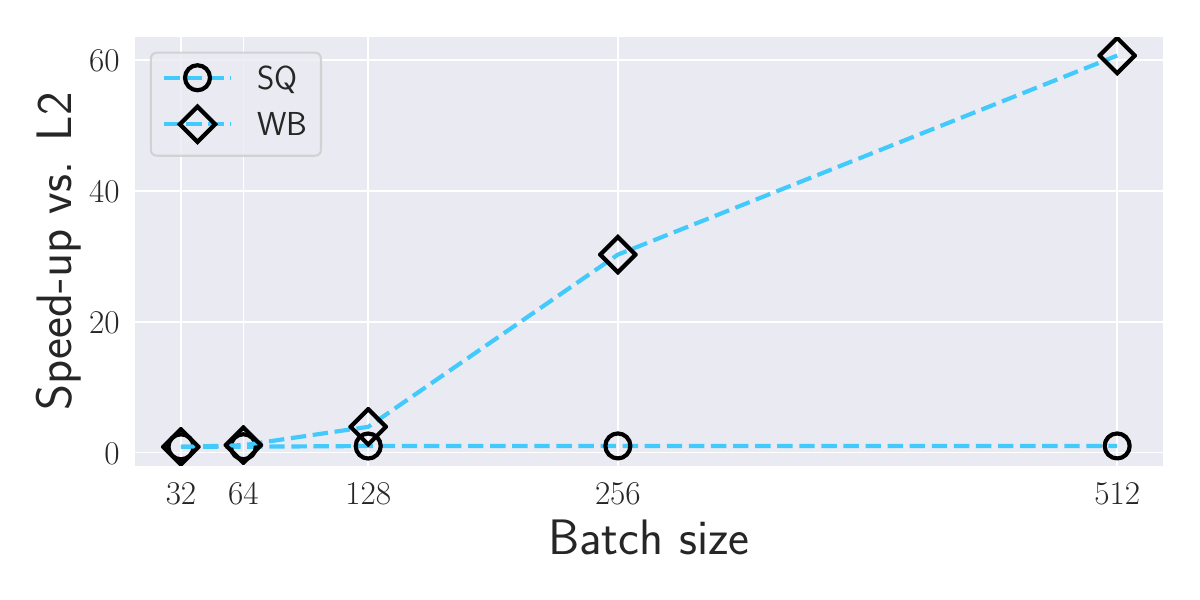}
    \end{subfigure}
    \begin{subfigure}[c]{.45\textwidth}
    \centering
    \includegraphics[width=1\textwidth]{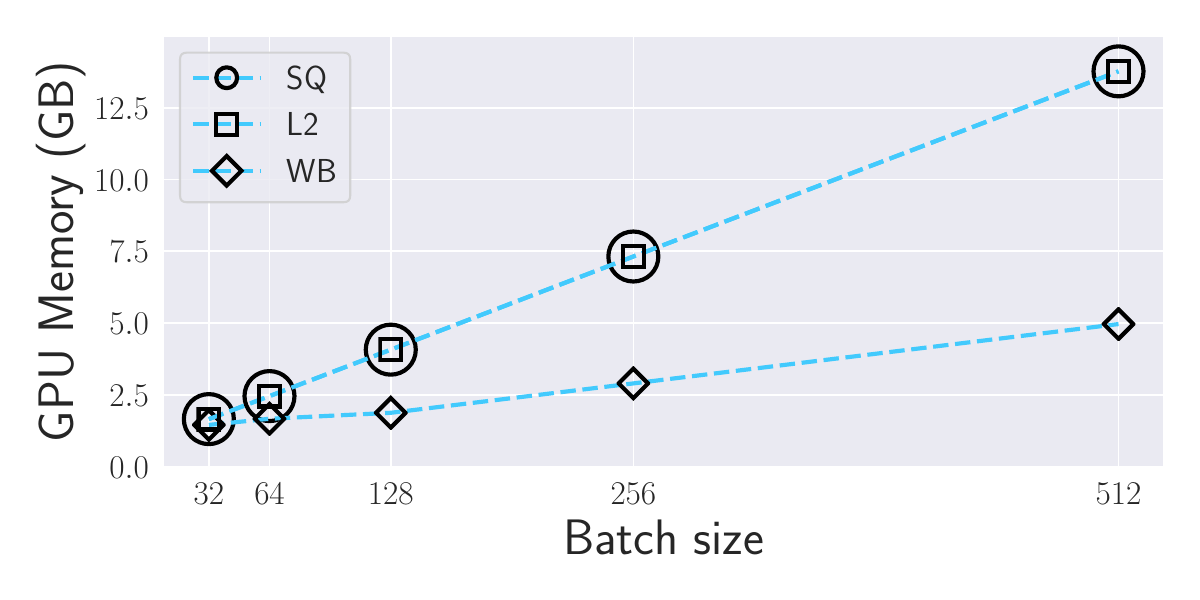}
    \end{subfigure}
    \caption{\small We show the speed-up per forward pass of the models with WB and SQ rescaling compared to L2 (\textbf{left}) as well as the GPU memory consumption (\textbf{right}) of the three different models in Table~\ref{tbl:result_table} for different batch sizes for both measures.
    While the models perform similarly for small batch sizes, the WB-based model scales better to large inputs.}
    \label{fig:nonlin_comparison_eff}
\end{figure*}
While all three non-linearities can yield interpretable CoDA Nets, the computational cost of the different approaches for bounding the DAU outputs differs. For example, by avoiding the explicit calculation of the $d$-dimensional weight vector in eq.~\eqref{eq:eDAU}, the eDAUs are able to save both memory as well as floating point operations---the computed vectors are of size $r\ll d$ and the dot-product in the low-dimensional space requires $O(r)$ operations instead of $O(d)$.
Being the fundamental building block of the CoDA Nets, such gains in efficiency can have considerable impact, since the corresponding computations are performed in every layer for every unit and at each spatial position of the input to the respective layer. Accordingly, in practice\footnote{In our experiments, we rely on the highly optimised implementations for convolutions from the pytorch \cite{pytorch} library.} we observed that the weight bounding approach in the eDAUs (eq.~\eqref{eq:eDAU}) can yield significant speed-ups and memory savings, especially for high-dimensional inputs. For example, in Fig.~\ref{fig:nonlin_comparison_eff} we plot the memory consumption and forward-pass speeds for the three different models without learnt embedding function (see Table~\ref{tbl:result_table}) for varying batch sizes on the CIFAR-10 dataset: while SQ and L2 perform similarly, the WB-based model scales better to larger inputs. 

Additionally, we measured memory consumption and training time for two models with the same architecture on Imagenet (L-CoDA, see beginning of sec.~\ref{sec:results}) for the SQ and the WB rescaling methods.
For this, we updated the models $\approx8000$ times with a batch size of 16 and recorded the overall time as well as the GPU memory consumption. In these experiments, the WB-based model required more than $3\times$ less memory (9.7GB vs.~30.0GB) and completed the updates more than $1.5\times$ faster (8.7 minutes vs.~14.1 minutes). All experiments regarding evaluation speed were performed on an nvidia Quadro RTX 8000 GPU with 48GB of memory.

\subsection{Hybrid CoDA Networks}
\label{subsec:interpolation_results}
In this section, we assess the interpretability of hybrid CoDA Nets, which combine conventional CNN layers and CoDA layers in one network model. For our experiments, we use varying numbers of pre-trained CNN layers as feature extractors on top of which a CoDA Net is trained as a classifier. Such a hybrid structure can prove useful in cases where CoDA Nets do not (yet) yield the same accuracy as conventional architectures; we kindly refer the reader to the supplement for details on the network architectures used in these experiments. 

In particular, we use the first K layers of a pre-trained ResNet model~\cite{he2016deep} as feature extractors. Since ResNets are piece-wise linear models, the hybrids are still \emph{dynamic linear} and we can assign importance values to input features according to their effective linear contribution; importantly, the input features can be extracted \emph{at any depth} of the network as the output of a CoDA layer or a ResNet block, or as the actual input pixels. In order to assess whether such hybrids are more interpretable than the base model, we compute spatial contribution maps\footnote{The contribution maps according to the dynamic linear mapping can be obtained via 'Input$\times$Gradient', where for the gradient calculation we treat the dynamic matrices in the CoDA layers as fixed.} with respect to different activation maps within the network and evaluate them under the localisation metric (see sec.~\ref{subsec:intp_results}, 'Evaluation metrics'). 

\begin{figure*}
\centering
    \begin{subfigure}[b]{0.48\textwidth}
    \centering
    \includegraphics[width=\textwidth]{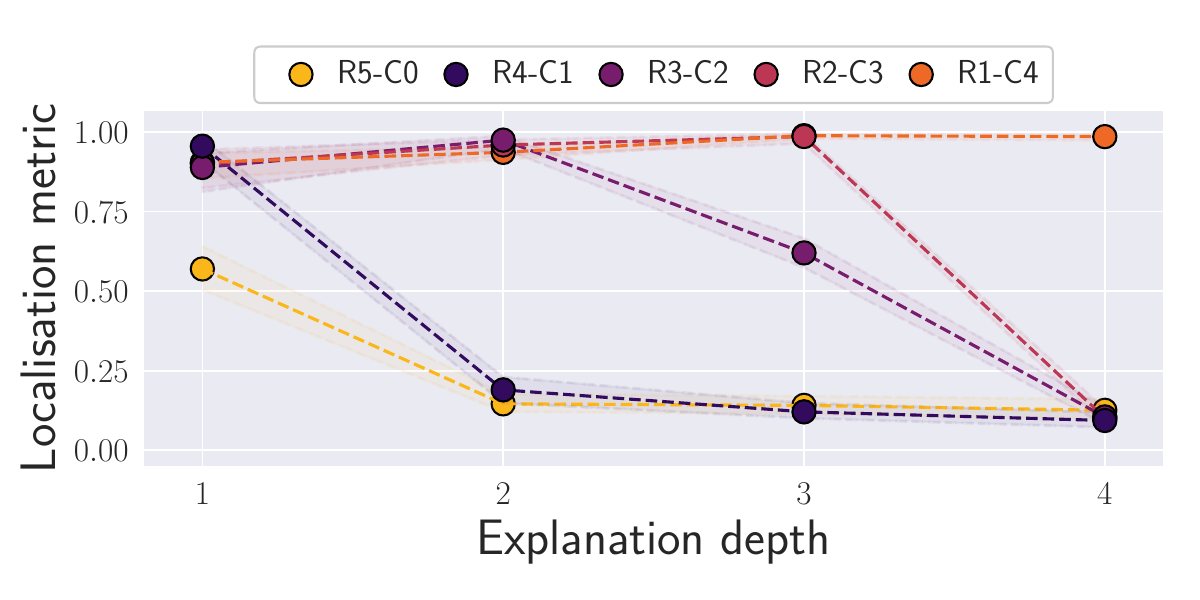}
    \end{subfigure}
    \begin{subfigure}[b]{0.48\textwidth}
    \centering
    \includegraphics[width=\textwidth]{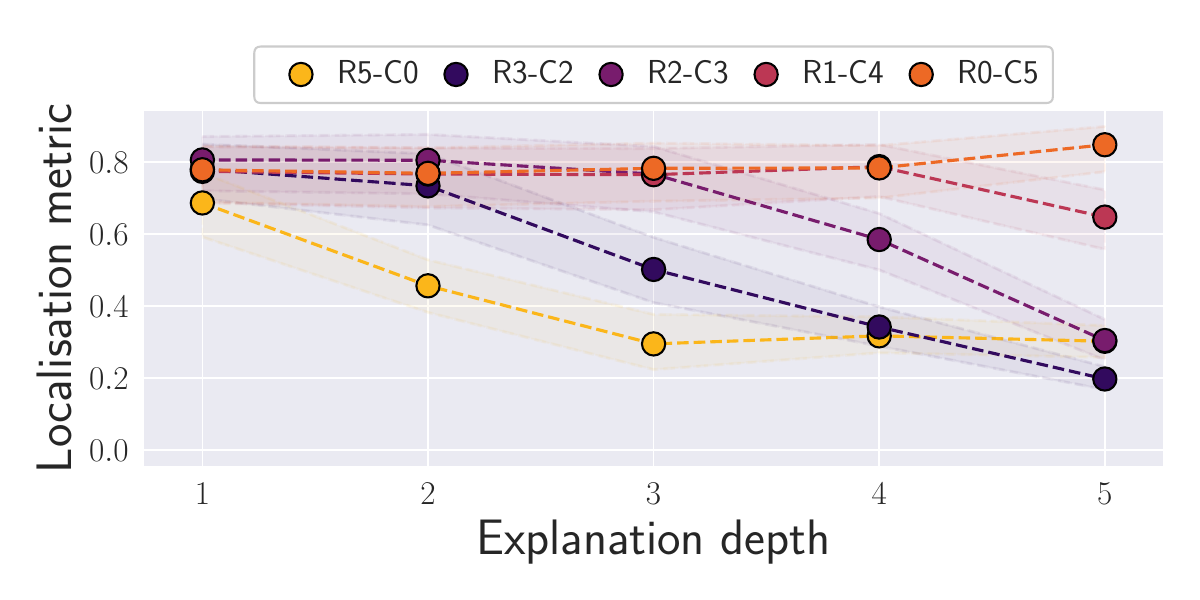}
    \end{subfigure}
    \caption{Localisation metric (mean and standard deviation) for different 'explanation depths' evaluated on four hybrid models trained on CIFAR-10 (\textbf{left}) and Imagenet (\textbf{right}). Additionally, we show the localisation results of the pretrained ResNets (denoted by R5-C0) that were used as base models. For each evaluation, we extract the effectively applied linear transformations up to a certain depth and compute the corresponding linear contributions to the output logits coming from individual positions in the activation maps. We then use the resulting  maps as an explanation of the output logit and assess how well these maps allow for localising the corresponding class images in the localisation task. As can be seen, the more layers of the original ResNet architectures are replaced by CoDA layers, the larger is the 'interpretable depth'.}
    \label{fig:interpolation}
\end{figure*}
\myparagraph{CIFAR10.} For the following experiments, we use a pre-trained ResNet-56 obtained from~\cite{Idelbayev18a}. This model consists of a convolutional layer + batch normalisation~\cite{ioffe2015batch} (C+B), followed by three times nine \emph{residual blocks} (RBs) as well as a fully connected and a pooling (FC+P) layer; for more details we kindly refer the reader to the original work~\cite{he2016deep} and the implementation~\cite{Idelbayev18a} on which we base these experiments. We can summarise this model by \mbox{[C+B, 9RB, 9RB, 9RB, FC+P]}; we will further denote individual \emph{segments} $\mathcal{S}_i$ of the model by their index in this summary counting from the back, e.g., $\mathcal{S}_5$=[C+B] and  $\mathcal{S}_1$=[FC+P]. In order to evaluate the interpretability of this model at different depths $t$, we split it at different points into two virtual parts: an embedding function $E_t(\vec x)$ and a classification head (CH$_t$). For a given split, we then regard the output of $E_t(\vec x)$  as the input to the classification head and linearly decompose the latter according to the respective linear transformation performed by the model; e.g., with an \emph{explanation depth} of $2$ we refer to the split in which $\mathcal{S}_2$ is the first element in CH$_2$ and
we evaluate linear contribution maps obtained for the classification head \mbox{CH$_2$=[9RB, FC+P]} on the preprocessed input \mbox{$E_2(\vec x)$=[C+B, 9RB, 9RB]$(\vec x)$}. By performing this evaluation for various splits, we can assess the 'interpretable depth' of a model. 
In particular, we evaluate how well the contribution maps at different depths allow for localising the correct class images in the localisation task.

In order to investigate the effect of CoDA layers on the interpretable depth, we train and evaluate four different hybrid models. For this, we replace an increasing number of segments $\mathcal S_i$ by CoDA layers, starting from $\mathcal S_1$; in Fig.~\ref{fig:interpolation} (\textbf{left}), we denote the base model by R5-C0 (5 ResNet segments, 0 CoDA segments) and the hybrids according to the number of replaced segments, e.g., for R3-C2 we replaced the last two segments by CoDA layers\footnote{In detail, each segment of 9RB is replaced by a set of 3 CoDA layers. The final network segment [FC+P] is replaced by a single CoDA layer followed by a global pooling operation.}.
\emph{For each of these models}, we can decompose the model outputs in terms of contributions from spatial positions for the embedding functions $E_t$ defined by different splits $t$. Note that the base model (ResNet-56) is piece-wise linear and we can thus still compute linear contributions at any depth of this hybrid network.
In Fig.~\ref{fig:interpolation} (\textbf{left}) we show the results of the localisation metric for all networks at various depths; the classification accuracies can be found in Table~\ref{tbl:imagenet}. As can be seen, the linear contributions are good explanations of the class logits as long as the classification head entirely consists of CoDA layers and drops as soon as we include a segment with ResNet blocks in the classification head CH. Again, this highlights that \emph{dynamic linearity} alone is not enough to obtain useful linear decompositions of the model outputs, but that the alignment property is crucial for the interpretability of the CoDA layers.

\myparagraph{Imagenet.} In the following, we show that the gains from interpolating between networks also extend to a more complex dataset.
Similar to the interpolation experiments on CIFAR-10 above, the results are based on an interpolation between a pretrained ResNet model (ResNet-50) and a CoDA-based classification head. However, given the high-dimensional representations produced by the later ResNet layers (up to 2048 channels), the parameters of the classification head increase drastically if the high dimensionality is maintained throughout the CoDA layers. Therefore, in the Imagenet experiments, we first compute a low-dimensional projection $\tilde{\vec x} = \mat P \vec x$ of the inputs to the convolutional kernels to which we apply the eDAUs (see eq.~\eqref{eq:eDAU}); similarly to the dynamic weights of DAUs with L2 normalisation, we normalise the rows of the matrix $\mat P$ to unit norm to maintain a parameter-independent bound of the network.

\begin{table}[!t]
\renewcommand{\arraystretch}{1.3}
\caption{Classification accuracies for hybrid networks. \mbox{R$X$-C$Y$} denotes how many segments were replaced by CoDA layers; on Imagenet, maximally up to five ResNet blocks from the end of the network were replaced and all networks thus still rely on a ResNet-based stem.
On CIFAR-10 the accuracy can be maintained whilst improving interpretability (see Fig.~\ref{fig:interpolation}). On Imagenet, on the other hand, we observe a trade-off in accuracy when increasing the 'interpretable depth' of the models.}
\label{tbl:imagenet}
\centering
\begin{tabular}{|c|c|c|c|c|c|}
\hline
\multirow{2}{*}{\bf CIFAR-10}&
R5-C0 & R4-C1 & R3-C2 & R2-C3 & R1-C4 \\
\cline{2-6}
&93.4\% & 93.6\% &  93.4\% &  93.6\% &   93.8\%  \\
\hline
\multirow{2}{*}{\bf Imagenet}& 
R5-C0 & R3-C2 & R2-C3 & R1-C4 & R0-C5 \\
\cline{2-6}
&76.1\% & 74.7\% &  73.3\% &  71.7\% &   71.4\% \\
\hline
\end{tabular}
\end{table}
For the interpolation experiments, we successively replace the last 5 residual blocks of the ResNet-50 base model (the 'segments' here correspond to FC+P or individual residual blocks) by a single CoDA layer each and assess the interpretability via the localisation metric as well as model accuracy, see Fig.~\ref{fig:interpolation} (\textbf{right}) and Table~\ref{tbl:imagenet} respectively. Similar to the CIFAR-10 experiments, we observe an increase in 'interpretable depth' (Fig.~\ref{fig:interpolation}, right). However, while on CIFAR-10 the accuracy of the base model could be maintained, on Imagenet we observe a trade-off in accuracy. While better results can certainly be achieved by further optimising the network architectures and / or fine-tuning the learnt embeddings, our results show that it is possible to increase the interpretability of performant classification models by using a classification head comprised of CoDA layers. 

\section{Discussion and conclusion}
\label{sec:discussion}
We present a new family of neural networks, the CoDA Nets, 
and show that they are performant classifiers with a high degree of interpretability\footnote{Code is available at \url{github.com/moboehle/CoDA-Nets}}.
In particular, we first introduced the Dynamic Alignment Units (DAUs), which model their output as a dynamic linear transformation of their input and have a structural bias towards alignment maximisation. This bias is induced by ensuring that a DAU can only produce large outputs if its weights are well-aligned with the input, since the dynamically applied weights are explicitly normalised. In order to lower the computational costs of the DAUs, we further introduce the eDAUs, for which we normalise the weights by an upper bound of their norms which is cheaper to compute.
    Using the DAUs to model filters in a convolutional network, we obtain the Convolutional Dynamic Alignment Networks (CoDA Nets).
The successive linear mappings by means of the DAUs within the network make it possible to linearly decompose the output into contributions from individual input dimensions---in contrast to piece-wise linear networks, which are also dynamic linear, the alignment property of the DAUs ensures that the linear decomposition aligns with discriminant patterns in the input.
In order to assess the quality of these contribution maps,
    see eq.~\eqref{eq:contrib}, we compare against other attribution methods.
    We find that the CoDA Net contribution maps consistently perform well under commonly used quantitative metrics and are robust to the applied normalisation scheme.
    Beyond their \emph{interpretability},
        the CoDA Nets constitute performant classifiers: their accuracy on CIFAR-10 and the TinyImagenet dataset are on par to the commonly employed VGG and ResNet models.
Lastly, we show that CoDA layers can be combined with conventional networks, which yields hybrid models with an increased 'interpretable depth' compared to the base model. We believe that such hybrid models hold great potential, since they take advantage of the high modelling capacity and efficiency of modern neural networks whilst allowing for a user-defined 'minimal interpretability'. For example, such networks could allow for localising regions of importance for the model decision at a desired granularity by restricting the receptive field of the feature extractors.




%
\bibliographystyle{IEEEtran}
\bibliography{biblio}

\newpage

%

\begin{IEEEbiography}[{\includegraphics[width=1in,height=1.25in,clip,keepaspectratio]{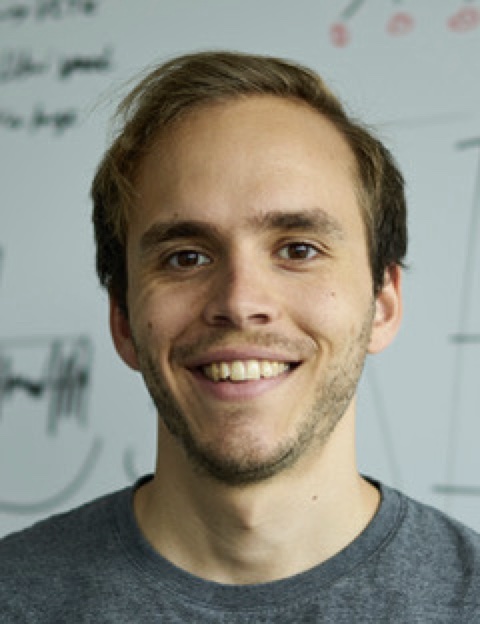}}]{Moritz Böhle}
is  a  Ph.D.~student  in  Computer Science at the Max Planck Institute for Informatics, working  with  Prof.  Dr.~Bernt  Schiele  and  Prof.~Dr.~Mario Fritz.  He graduated with a bachelor's degree in physics in 2016 from the Freie Universität Berlin and obtained his Master's degree in computational neuroscience in 2019 from the Technische Universität Berlin. His research focuses on understanding the 'decision process' in deep neural networks and designing inherently interpretable neural network models.
\end{IEEEbiography}

\begin{IEEEbiography}[{\includegraphics[width=1in, height=1.25in, trim=0 13.5em 6em 0, clip, keepaspectratio]{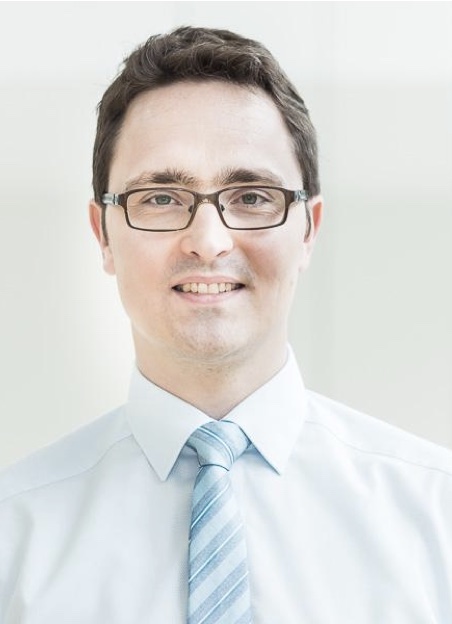}}]{Mario Fritz}
Mario Fritz is faculty member at the CISPA Helmholtz Center for Information Security, honorary professor at the Saarland University, and a fellow of the European Laboratory of Learning and Intelligent Systems (ELLIS). Before, he was senior researcher at the Max Planck Institute for Informatics, PostDoc at UC Berkeley and International Computer Science Institute. He studied computer science at the university Erlangen/Nuremberg and obtained his PhD from the TU Darmstadt. His current work is centered around Trustworthy Information Processing with a focus on the intersection of AI \& Machine Learning with Security \& Privacy. He is associate editor of IEEE TPAMI, a member of the ACM Europe Technical Policy Committee Europe, and a leading scientist of the Helmholtz Medical Security, Privacy, and AI Research Center, where he is coordinating projects on privacy and federated learning in health. He has over 100 publications, including 80 in top-tier journals (IJCV, TPAMI) and conferences (NeurIPS, AAAI, IJCAI, ICLR, NDSS, USENIX Security, CCS, S\&P, CVPR, ICCV, ECCV).
\end{IEEEbiography}


\begin{IEEEbiography}[{\includegraphics[width=1in,height=1.25in,clip,keepaspectratio]{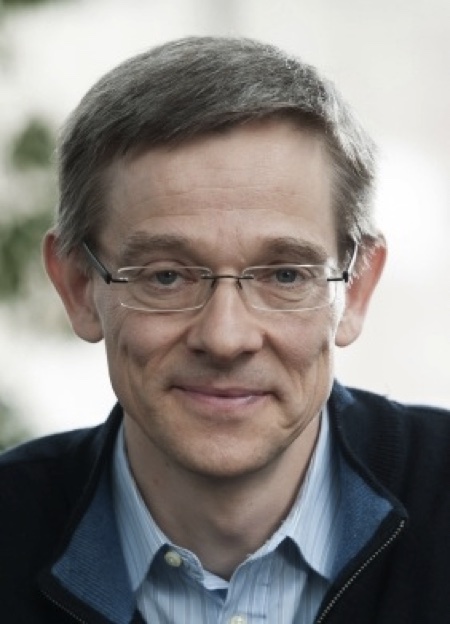}}]{Bernt Schiele}
has  been  Max  Planck  Director at  MPI  for  Informatics  and  Professor  at  Saarland  University  since  2010.  He  studied  computer  science  at  the  University  of  Karlsruhe, Germany.  He  worked  on  his  master  thesis  in the field of robotics in Grenoble, France, where he  also  obtained  the  ``diplome  d’etudes  approfondies d’informatique”. In 1994 he worked in the field of multi-modal human-computer interfaces at  Carnegie  Mellon  University,  Pittsburgh,  PA, USA  in  the  group  of  Alex  Waibel.  In  1997  he obtained his PhD from INP Grenoble, France under the supervision of Prof.~James L.~Crowley in the field of computer vision. The title of his thesis was ``Object Recognition using Multidimensional Receptive Field Histograms”. Between 1997 and 2000 he was postdoctoral associate and Visiting Assistant Professor with the group of Prof. Alex Pentland at the Media Laboratory of the Massachusetts Institute of Technology, Cambridge, MA, USA. From 1999 until 2004 he was Assistant Professor at  the  Swiss  Federal  Institute  of  Technology  in  Zurich  (ETH  Zurich). Between 2004 and 2010 he was Full Professor at the computer science department of TU Darmstadt.
\end{IEEEbiography}




\newpage


\section*{Supplementary material}

# Optimising for Interpretability:
# Convolutional Dynamic Alignment Networks

## Supplementary Material

Moritz Böhle
MPI for Informatics
Saarland Informatics Campus

Mario Fritz
CISPA Helmholtz Center
for Information Security

Bernt Schiele
MPI for Informatics
Saarland Informatics Campus

# Table of Contents

In this supplement to our work on Convolutional Dynamic Alignment Networks (CoDA-Nets), we provide:

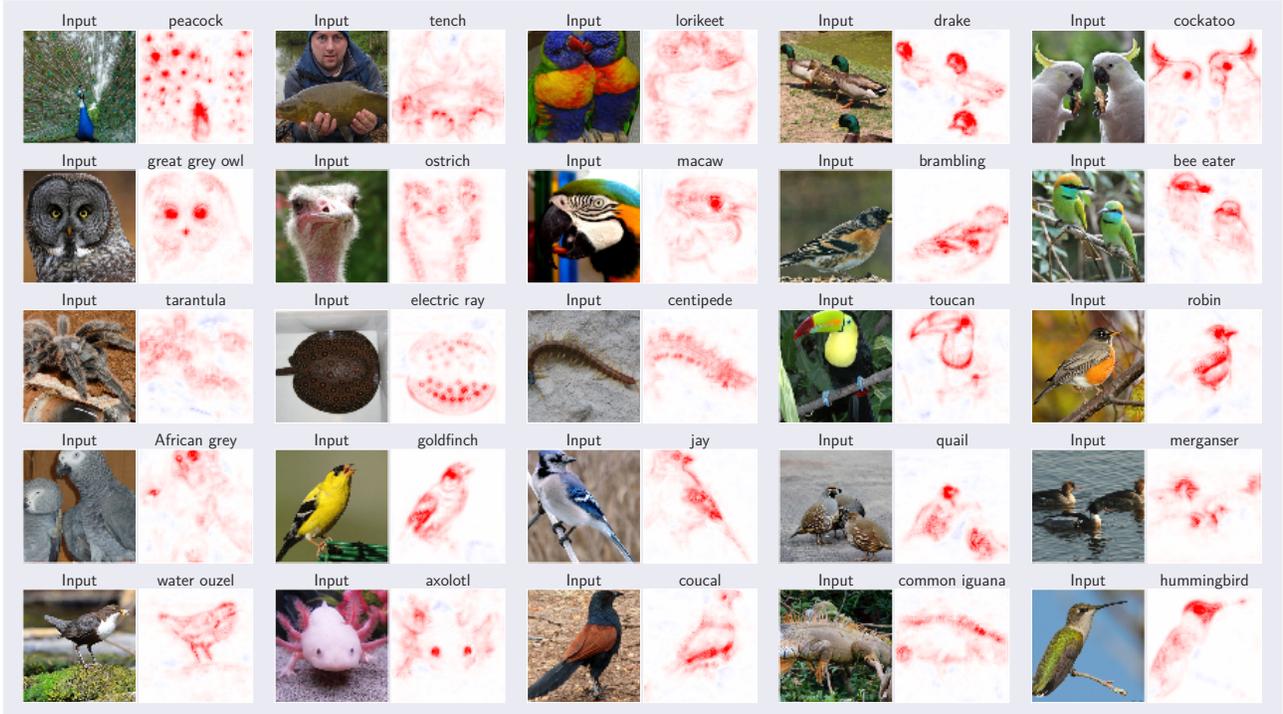

Figure A1: The first 25 most confident classifications decomposed into the contributions from each spatial location, filtered to 1 image per class. Positive (negative) contributions for the ground truth class are shown in red (blue).

## A. Additional qualitative results

**Additional Imagenet examples.** In Figs. A1 and A2 we present additional qualitative examples of the model-inherent contribution maps. In particular, we show the decomposition of the model predictions into input contributions for 50 out of 100 classes; for each class, we show the most confidently classified image and show the classes in sorted order (by confidence). In Figs. A3 and A4, we moreover show the contributions maps for the first 10 of the overall most confidently classified images next to the attribution maps from the post-hoc importance attribution methods for qualitative comparison. We note that GradCam consistently highlights very similar regions to the CoDA-Net contribution maps, but does so at a lower resolution. All contribution maps based on the CoDA-Net use the same linear color scale, which has been set to $(-v, v)$ with $v$ the 99.75th percentile over all absolute values in the contributions maps shown in Figs. A1 and A2. For reproducing the presented contribution maps and more, please visit github.com/moboehle/CoDA-Nets.

**Regularising the linear mapping on CIFAR-10.** As mentioned in the main paper, the explicit representation of the model computations as a linear mapping, i.e.,

$$\hat{\mathbf{y}}(\mathbf{x}) = \mathbf{W}_{0 \to L}(\mathbf{x})\mathbf{x} \quad,$$

allows to explicitly regularise the linear mappings and thereby the model-inherent contribution maps. In Fig. A5 we qualitatively show how the regularisation impacts the contribution maps; for these experiments, we added a regularisation term to the loss function and optimised

$$\mathcal{L}(\mathbf{x}_i, \mathbf{y}_i) = \text{BCE}\left(\sigma(T^{-1}\mathbf{W}_{0 \to L}(\mathbf{x}_i)\mathbf{x}_i + \mathbf{b}_0), \mathbf{y}_i\right) + \lambda \langle |\mathbf{W}_{0 \to L}(\mathbf{x}_i)| \rangle \tag{A.1}$$

with $\langle |\mathbf{U}| \rangle$ denoting the average absolute value of a matrix $\mathbf{U}$. We see that without any regularisation the contribution maps are difficult to interpret. As soon as even a small regularisation is applied, however, the maps align well with the discriminative parts of the input image.

**Sanity check.** In [S1], the authors found that many commonly used methods for importance attribution are not *model-faithful*, i.e., they do not reflect the learnt parameters of the model. In order to test this, they proposed to examine how the attributions change if the model parameters are randomised step by step. If the attributions remain stable under model randomisation, they cannot be assumed to explain a specific model, but rather reflect properties of the general architecture

15

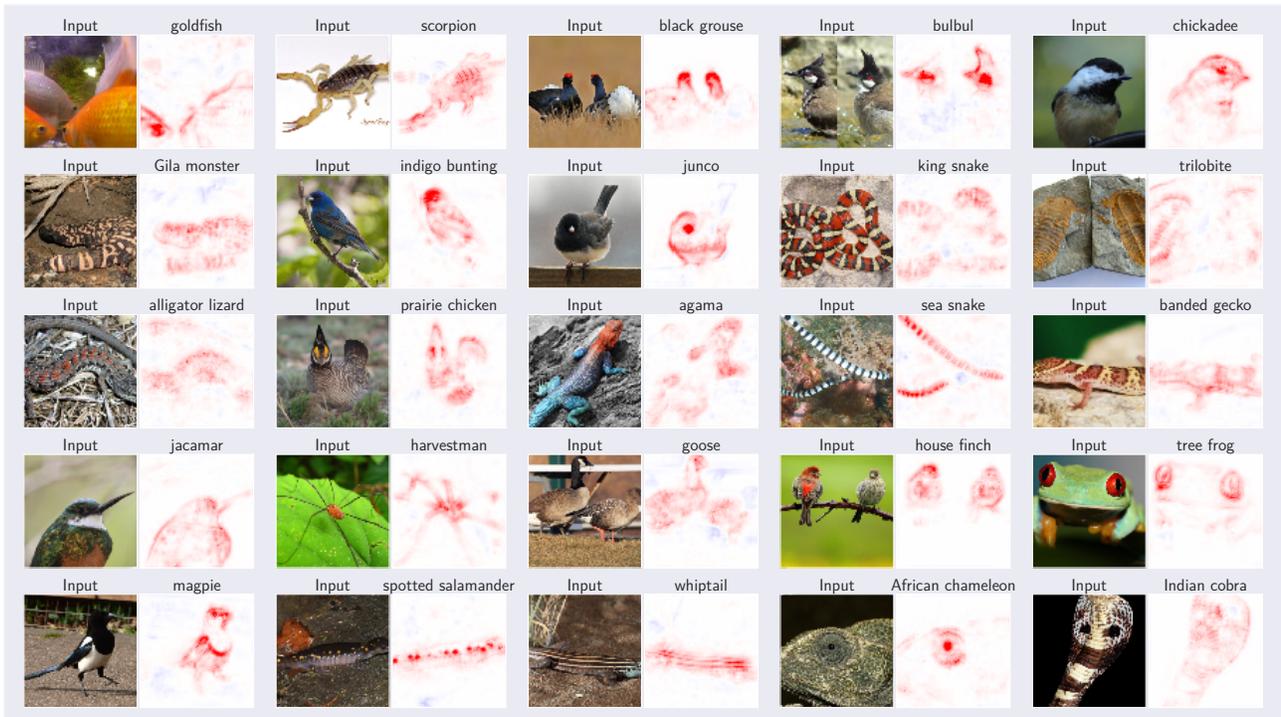

Figure A2: The 26th to the 50th most confident classifications decomposed into the contributions from each spatial location, filtered to 1 image per class. Positive (negative) contributions for the ground truth class are shown in red (blue).

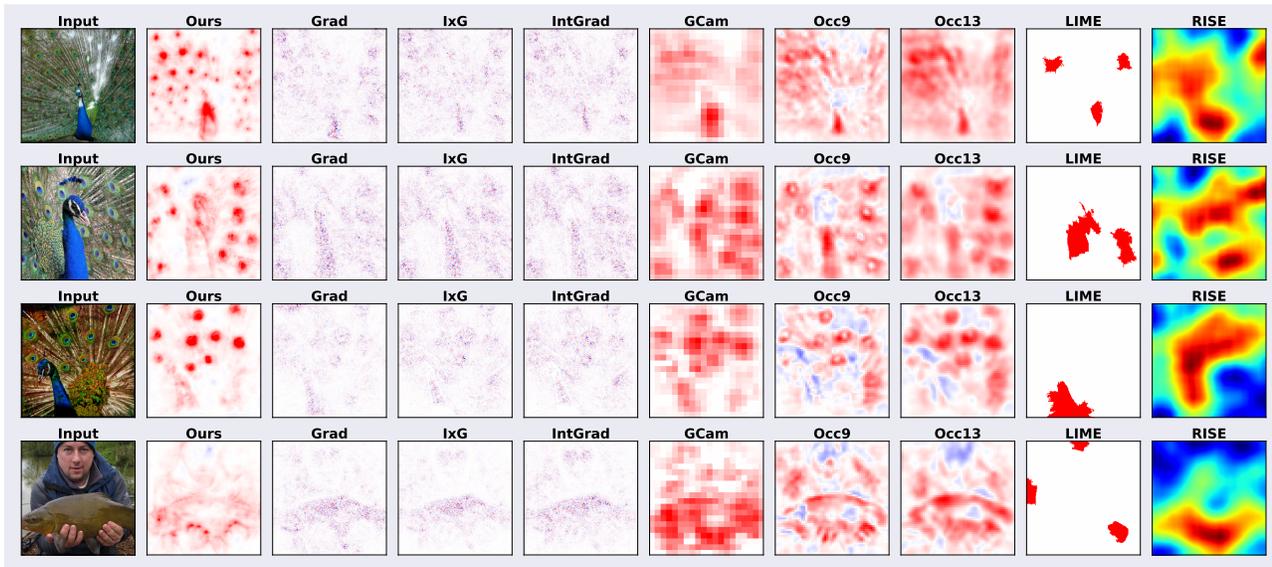

Figure A3: Comparison between attribution methods for the 4 most confident classifications. We show positive importance attributions in red, negative attributions in blue; for RISE we use its default visualisation. Note that the model seems to align the weights well with the ornamental eyespots of the peacocks (also see the peacock in Fig. A4) or the heads of the ducks in Fig. A4. While the latter are also highlighted by GCam, the former constitute a structure that is too fine-grained for GCam to resolve properly.

and the input data. In Fig. A6, we show how the model-inherent contribution maps behave when re-initialising the CoDA-Net

16

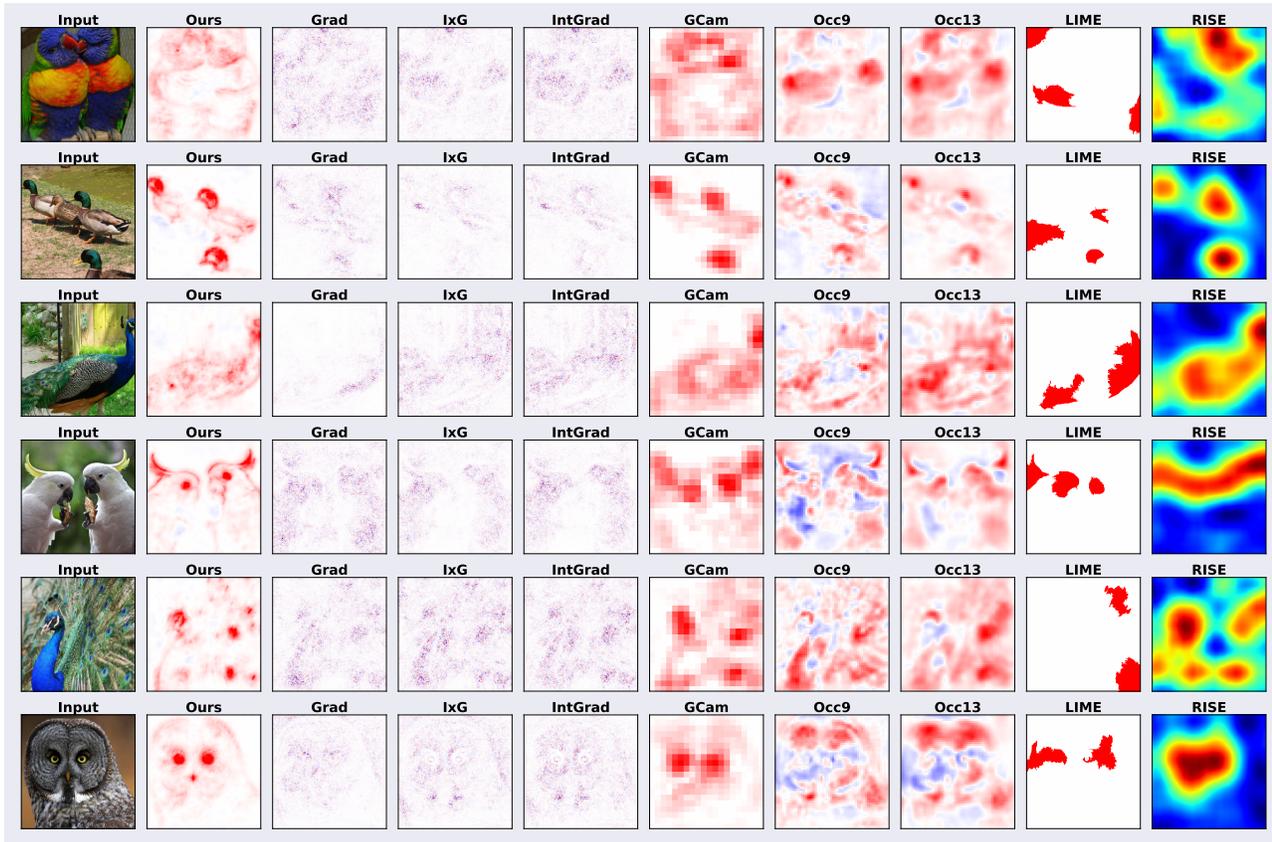

Figure A4: Comparison between attribution methods for the 5th to the 10th most confident classifications. We show positive importance attributions in red, negative attributions in blue; for RISE we use its default visualisation.

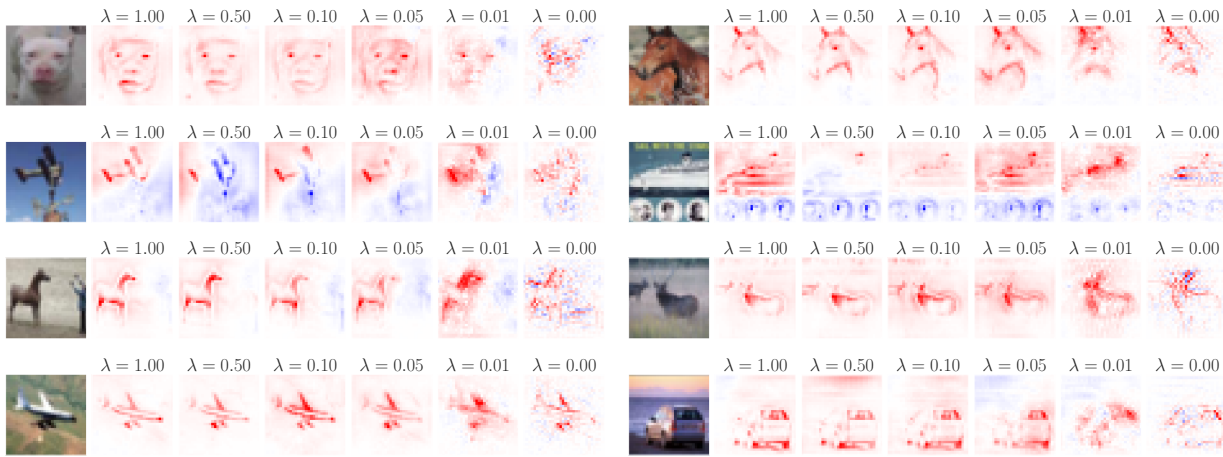

Figure A5: Visualising the qualitative effect of explicitly regularising the linear mapping $\mathbf{W}_{0 \to L}$ on CIFAR-10. While the contribution maps without regularisation are noisy, they become sharper as soon as a regularisation is applied.

layers one at a time to a random parameter setting, starting from the deepest layer. As can be seen, the contribution maps get significantly perturbed with every layer that is reset to random parameters; thus, the contribution maps pass this 'sanity check' for attribution methods.

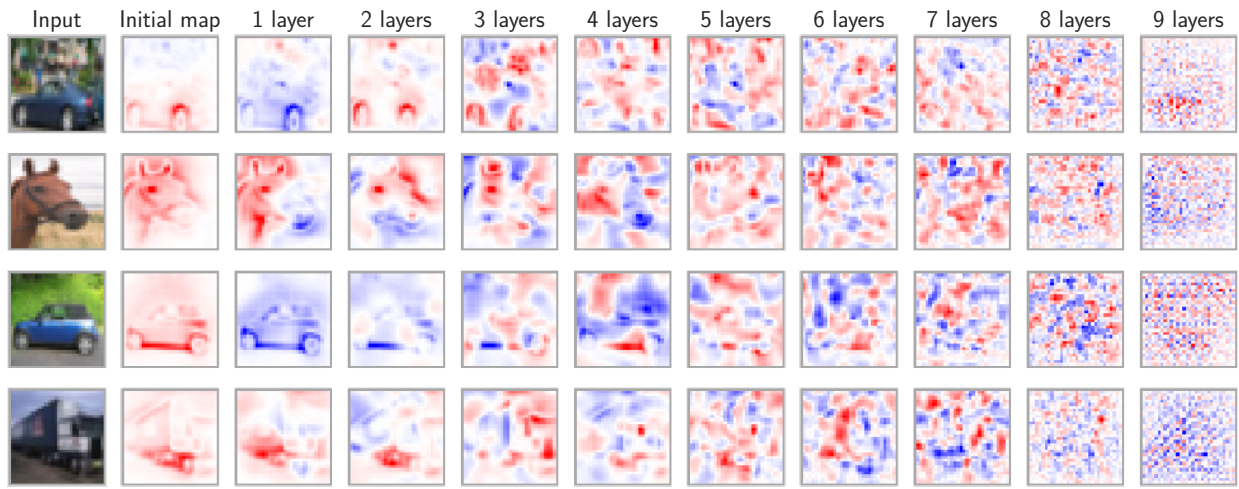

Figure A6: Sanity check experiment as in [S1]. If the importance attributions remain stable under parameter randomisation, they cannot be assumed to faithfully reflect the learnt parameters of the model. Since the contribution maps get significantly perturbed when re-initialising layers from network output (left) to network input (right), the model-inherent contribution maps thus pass this sanity check. Positive contributions are shown in red, negative contributions in blue.

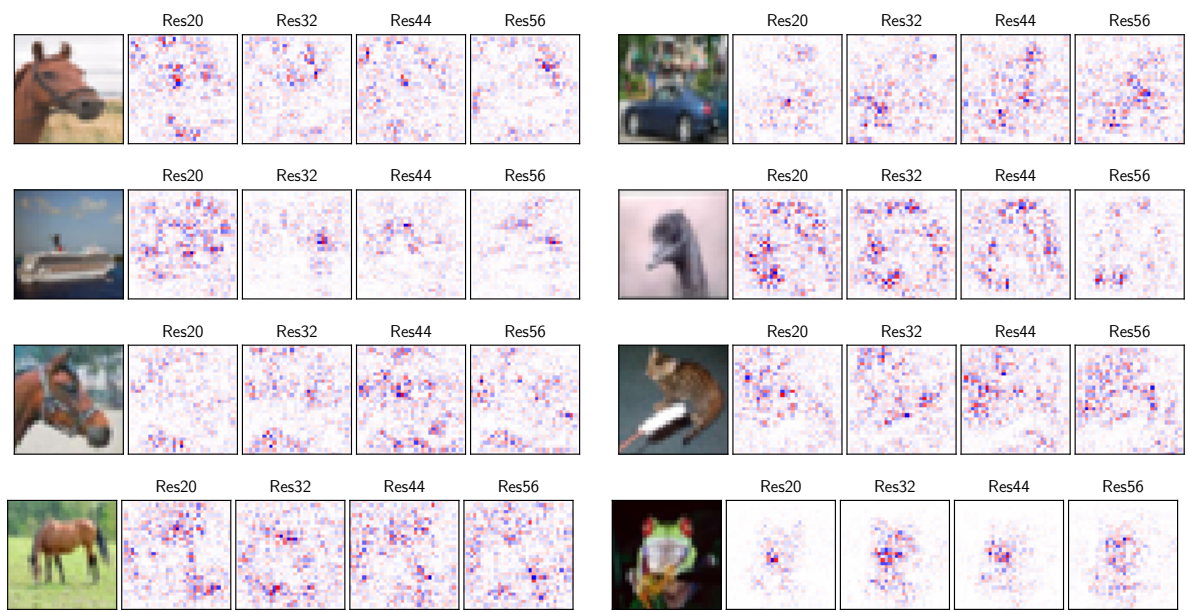

Figure A7: 'Input×Gradient' evaluated on different ResNets. Since ResNets are piece-wise linear functions, s.t. $\mathbf{y}(\mathbf{x}) = \mathbf{M}(\mathbf{x})\mathbf{x} + \mathbf{b}(\mathbf{x})$, this is the ResNet-based equivalent to the CoDA-Net contribution maps.

**Contribution maps of a piece-wise linear model.** In Fig. A7 we show contribution maps obtained from different pre-trained ResNet architectures obtained from https://github.com/akamaster/pytorch_resnet_cifar10. In particular, we visualise the 'Input×Gradient' method. Note that this yields contribution maps, since piece-wise linear models, such as the ResNets, produce input-dependent linear mappings, similar to the CoDA-Nets. These contribution maps, however, are rather noisy and do not reveal particularly relevant features.

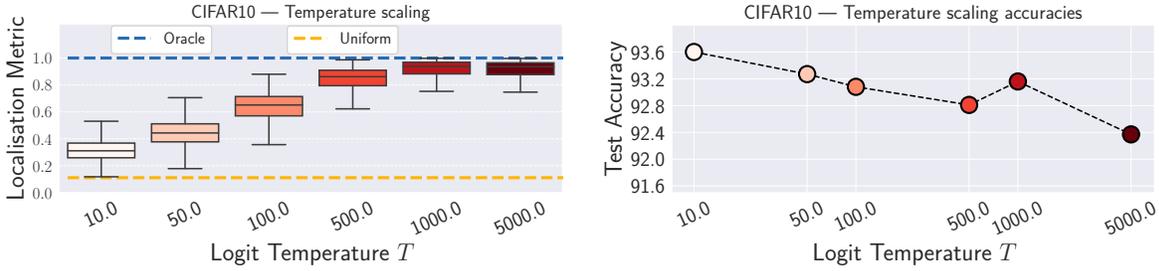

Figure B1: **Left:** Localisation metric results for models trained with different temperatures $T$, same as in main paper (see Fig. 6, top right). **Right:** Corresponding accuracies of the models on the CIFAR10 test set. There seems to be a trade-off between the interpretability and the accuracy of the models due to the regularising effect of $T$.

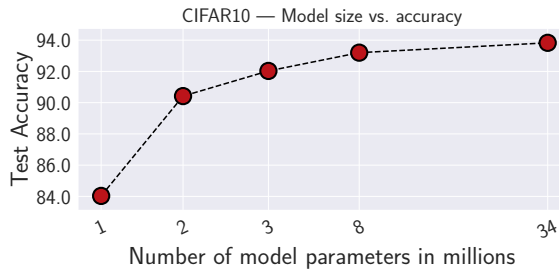

Figure B2: Effect of scaling the rank of the DAUs in the CoDA-Nets on accuracy. Specifically, all ranks in the S-CoDA-SQ model presented in the main paper were scaled with the same factor, thereby changing the model size.

## B. Additional quantitative results

**Performance / interpretability trade-off.** While the CoDA-Nets were observed to train and perform well over a wide range of choices for the logit temperature $T$, there seems to be a trade-off between the accuracies of the network and their interpretability—the implicit alignment regularisation comes at a cost. For example, in Fig. B1, we contrast the gain in interpretability (left, same figure as in main paper) with the corresponding accuracies (right).

**Model size vs. accuracy.** Given the quadratic form in the DAUs (cf. eq. (1), $\approx \mathbf{x}^T\mathbf{M}\mathbf{x}$), the number of parameters per DAU scales quadratically with the input dimensions. In order to limit the model size, we decided to explicitly limit the rank of the DAUs by factorising the matrix $\mathbf{M}$ into $\mathbf{AB}$ with $\mathbf{A} \in \mathbb{R}^{d\times r}$ and $\mathbf{B} \in \mathbb{R}^{r\times d}$. While this allows to be more parameter efficient, it, of course, affects the modelling capacity of the DAUs. In Fig. B2, we present how the accuracy changes with the model size; for this, we scaled the ranks of all DAUs per layer with factors of 1/8, 1/4, 1/2, and 4.5 compared to the S-CoDA-SQ model presented in the main paper. This results in models with 1.1M, 2.0M, 4.0M, and 34.6M parameters respectively. For comparison, the original model is also included in Fig. B2 (8M parameters).

**Interpretability results for L2 and SQ non-linearity.** In Fig. B3a we show the results of evaluating the different methods for importance attribution on a model with the L2 and SQ non-linearities, see eq. (2) in the main paper. As can be seen, the results are very similar to those presented in the main paper in Fig. 6 (center column) for a model trained with the WB rescaling method; in particular, the model-inherent contribution maps outperform the other methods under the localisation metric and are on par with the occlusion methods under the pixel removal metric; note, however, that the occlusion methods are a direct estimate of the behaviour under pixel removal and therefore expected to perform well under this metric.

**Regularisation of $\mathbf{W}_{0\to L}$.** In Fig. B3b, we show the results of the localisation metric and the pixel removal metric for models trained on CIFAR-10 with different regularisation strengths $\lambda$, see eq. (A.1). We note that the localisation metric benefits from an increase in the regularisation strength $\lambda$. For the pixel removal metric, we observe that while the models improve at first under this metric, a strong regularisation makes the predictions more brittle.

**Removing the *most important* pixels first.** In Fig. B4a, we show the results of the pixel removal metric when removing the most important pixels first. As can be seen, the model-inherent contributions better predict the importance of pixels and

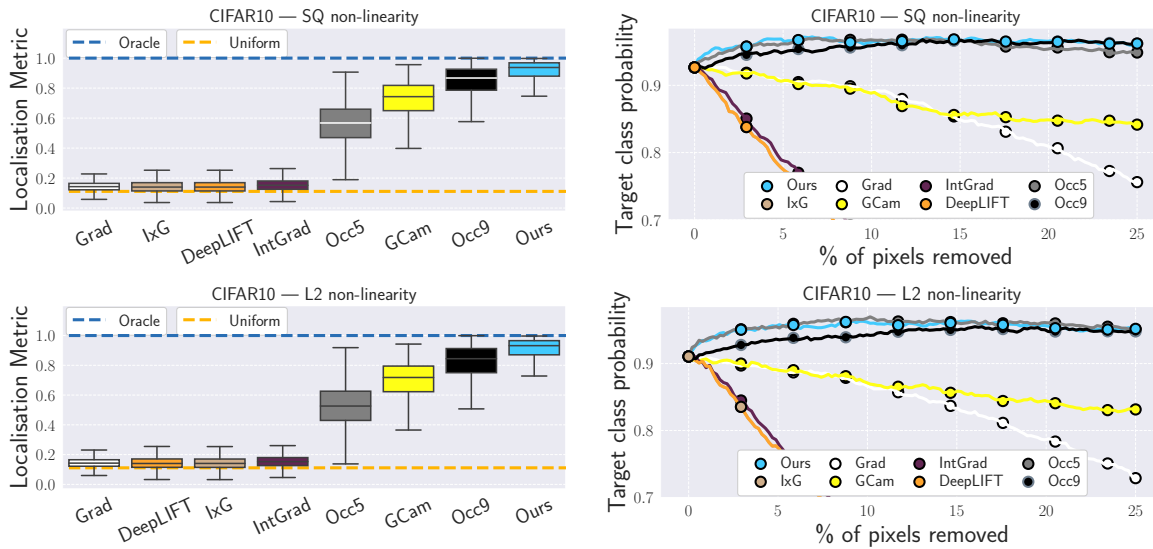

(a) The quantitative results for CoDA-Nets trained with the L2 and SQ non-linearities are very similar to those of a model trained with the WB non-linearity as shown in the main paper (Fig. 6, center column). In particular, we observe that the CoDA-Net outperforms the other methods under the localisation metric and achieves similar performance to the occlusion attribution method, which directly estimates the change in model prediction when removing a pixel.

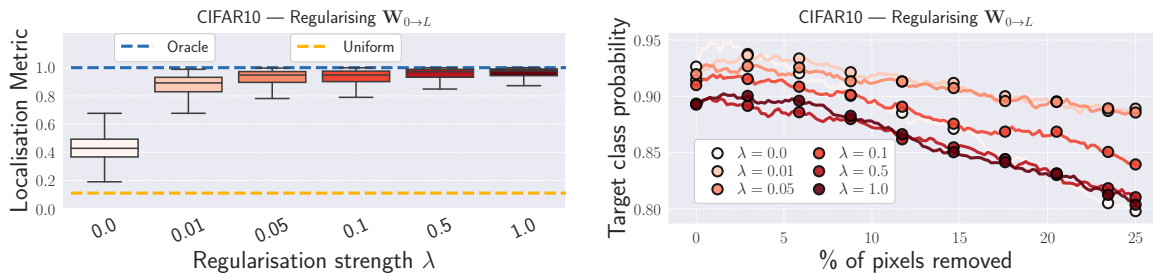

(b) Similar to the effect of temperature scaling, cf. Fig. 6 (right column) in the main paper, here we show quantitative results for different regularisation strengths $\lambda$, see eq. (A.1). Similar to increasing the temperature, stronger regularisations also improve the localisation metric (left). While the pixel perturbation metric (right) also improves at first ($\lambda = 0.01$ and $\lambda = 0.05$), the models become less stable for stronger regularisation strengths. In comparison, we found that increasing the temperature also improves the performance of the models under this metric, see Fig. 6 (right column) in the main paper.

Figure B3: Quantitative results for two ablations: (a) using the L2 or SQ non-linearities and (b) increasing the regularisation of the linear mapping $\mathbf{W}_{0 \rightarrow L}$, see eq. (2) in the main paper and eq. (A.1) respectively. Compare with Fig. 6 in the main paper.

outperform the other attribution methods under this metric; in particular, the confidence drops most rapidly when removing pixels according to the ranking given by the model-inherent contribution maps.

**Evaluating a pre-trained ResNet.** In order to establish a baseline for the performance of the different attribution methods on a classical neural network architecture, we additionally evaluated the attribution methods on a pre-trained ResNet and show the results in Fig. B4b. Specifically, we rely on a publicly available pre-trained ResNet-56 obtained from https://github.com/akamaster/pytorch_resnet_cifar10, which achieves a classification accuracy of 93.39%. The results show that the performance of the CoDA-Net-derived contribution maps is not only strong when comparing them to attribution methods evaluated *on the same model*. Instead, they also perform well in comparison to those methods evaluated *on a different model*. In particular, the model-inherent contribution maps of the CoDA-Net outperform attribution methods evaluated on the pre-trained ResNet-56 under the localisation metric. Further, only the occlusion attributions produce similarly strong pixel importance rankings for the ResNet; note, however, that the occlusion methods are a

20

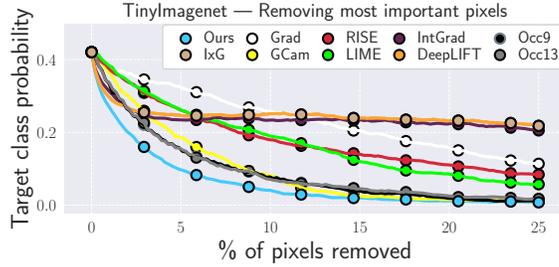

(a) Results for the pixel removal metric when first removing the most important pixels, evaluated on the S-CoDA-SQ model. As can be seen, the ranking given by the model-inherent contribution maps seems to best reflect the pixel importance, since the confidence most rapidly drops when removing pixels according to this ranking.

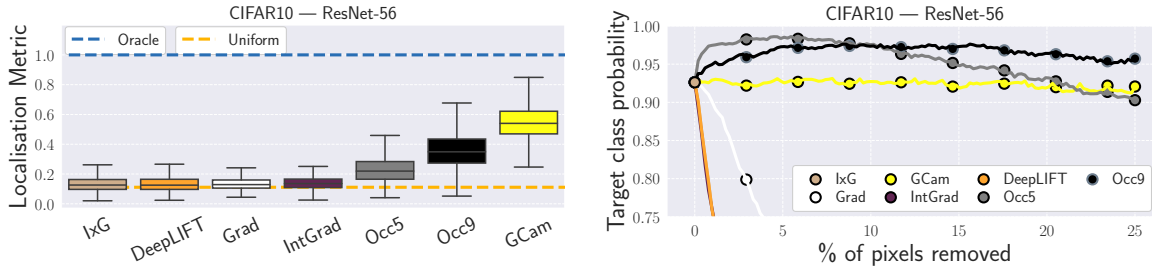

(b) Quantitative results for methods for importance attribtution evaluated on a pre-trained ResNet-56 on CIFAR-10. Comparing these results to Fig. 6 (center column) in the main paper or Fig. B3a, it can be seen that the model-inherent contribution maps of the CoDA-Net are also strong when compared to importance attributions evaluated on a different model.

Figure B4: In (a) we show the results of the pixel-removal metric when removing those pixels first that are considered the most important ones according to the importance attribution method. Moreover, in (b) we plot the quantitative results for the evaluation metrics of the importance attributions for a pre-trained ResNet-56.

direct estimate of the behaviour under pixel removal and therefore expected to perform well under this metric.

## C. Implementation details

### C.1. Training and architecture details

#### C.1.1 Pure CoDA Networks

**Architectures.** The architectures used for the results in Table 1 in the main paper are given in Table C1. All activation maps are padded with $(k-1)/2$ zeros on each side, such that the spatial dimensions are only reduced by the strides; here, $k$ refers to the kernel size. As can be seen, the activation maps thus still have a spatial resolution after the last layer, which we further reduce with a global sum-pooling layer. Note that global sum-pooling is just a linear layer with no trainable parameters and therefore still allows for linear decomposition. For the models which use a static embedding (the image and its negative, see sec. 4.3), the input itself consists of 6 channels; hence, the first layer takes an input with 6 channels per pixel. For the models with a learnt embedding function, the input has 32 channels. Note that for the eCoDA model the matrices $\mathbf{B}$ are not shared and each DAU has its own matrix $\mathbf{B}$.

**Training details.** We use the pytorch library [S9] and optimise all networks with the Adam optimiser [S8] with default values. As for the loss function, we use the binary cross entropy loss to optimise class probabilities individually (as 'one-vs-all' classifiers). For all networks, we used a base learning rate of $2.5 \times 10^{-4}$; for the Imagenet experiment and the S-eCoDA experiments on CIFAR-10, we employed learning rate warm-up and linearly increased the learning rate from $2.5 \times 10^{-4}$ to $1 \times 10^{-3}$ over the first 15 (10) epochs. Further, we trained for 200 epochs on CIFAR-10, for 100 epochs on TinyImagenet, and for 60 epochs on the Imagenet subset; we decreased the learning rate by a factor of 2 after every 60/30/20 epochs on CIFAR-10/TinyImagenet/Imagenet; for the eCoDA experiments on CIFAR-10 the learning rate was decayed according to a cosine schedule to a final learning rate of $10^{-5}$ after the initial linear increase. We used a batch size of 16, 128, and 64 for CIFAR-

21

10, TinyImagenet, and Imagenet respectively (and 64 for e-CoDA experiments on CIFAR-10). For the Imagenet subset, we additionally used RandAugment [S3] with parameters $n = 2$ and $m = 9$; for this, we relied on the publicly available implementation at https://github.com/ildoonet/pytorch-randaugment and followed their augmentation scheme. The qualitatively evaluated model (see Figs. 5, 8, and A1-A4) for the Imagenet subset was trained with $T = 1e5$ and achieved a top-1 accuracy of 76.5%. For comparison, we trained several ResNet-50 models (taken from the pytorch library [S9]) with the exact same training procedure, i.e., batch size, learning rate, optimiser, augmentation, etc.). The best ResNet-50 out of 4 runs achieved 79.16% top-1 accuracy[1], which outperforms the CoDA-Net but is nevertheless comparable. While it is surely possible to achieve better accuracies for both models, long training times for the CoDA-Nets have thus far prevented us from properly optimising the architectures both on the 100 classes subset, as well as on the full Imagenet dataset. In order to scale the CoDA-Net models to larger datasets, we believe it is important to first improve the model efficiency in future work. Lastly, when regularising the matrix entries of $\mathbf{W}_{0 \to L}$, see eq. (A.1), we regularised the absolute values for the true class $c$, $[\mathbf{M}_{0 \to L}]_c$, and a randomly sampled incorrect class per image.

### C.1.2 Hybrid CoDA Nets

**CIFAR10.** As described in the main paper, the interpolation experiments for CIFAR-10 are based on a ResNet-56 obtained from [S6]. This model consists of a convolutional layer + batch normalisation [S7] (C+B), followed by three times nine *residual blocks* (RBs) as well as a fully connected and a pooling (FC+P) layer; for more details we kindly refer the reader to the original work [S5] and the implementation [S6] on which we base these experiments. We can summarise this model by [C+B, 9RB, 9RB, 9RB, FC+P] and denote individual *segments* $\mathcal{S}_i$ of the model by their index in this summary counting from the back, e.g., $\mathcal{S}_5$=[C+B] and $\mathcal{S}_1$=[FC+P]. Similarly, we divide the S-eCoDA (see Table C1) into segments, which we define as [C+B, 3eCoDA, 3eCoDA, 2eCoDA, eCoDA+P]. We 'interpolate' between these two networks by successively replacing segments from the ResNet model by its corresponding segment in the CoDA Net—each replacement yields one hybrid model, which we train on the fixed representations[2] obtained from the ResNet-based stem. We use the same training scheme as in the S-eCoDA-based models on CIFAR-10 without learning rate warm-up. Further, the models with $[1, 2, 3, 4]$ replaced segments are trained with temperatures of $10^{[1,3,5,5]}$.

**Imagenet.** The interpolation experiments on the Imagenet dataset are based on a pre-trained ResNet-50 obtained from the pytorch [S9] library. Here, we define individual segments by the final classification layer + pooling and individual ResNet bottlneck blocks. For the corresponding eCoDA Net, the segments contain exactly one eCoDA layer. We train with a batch size of 128, DAU ranks of 8, 100 epochs, a cosine learning rate decay from $10^{-3}$ to $10^{-5}$, a temperature $T = 100$ and apply RandAugment [S3] with default parameters on random crops of size $192 \times 192$.

### C.2. Convolutional Dynamic Alignment Units

In Algorithm 1, we present the implementation of the convolutional DAUs (CoDAUs). As can be seen, a Convolutional Dynamic Alignment Layer applies dynamic (input-dependent) filters to each of the patches extracted at different spatial locations. In detail, the Dynamic Alignment Units are implemented as two consecutive convolutions (lines 11 and 15), which are equivalent to first applying matrix $\mathbf{B}$ (line 24) and then $\mathbf{A}$ to each patch and adding a bias term $\mathbf{b}$ (line 29). After applying the non-linearity (line 33), we obtain the dynamic weight vectors for CoDAUs as described in eq. (1) in the main paper. In particular, for every patch $\mathbf{p}_{hw}$ extracted at the spatial positions $hw$ in the input, we obtain the dynamic weight $\mathbf{w}_j(\mathbf{p}_{hw})$ for the $j$-th DAU as

$$\mathbf{w}_j(\mathbf{p}_{hw}) = g(\mathbf{A}_j \mathbf{B} \mathbf{p}_{hw} + \mathbf{b}_j) \; ; \tag{C.1}$$

note that the projection matrices $\mathbf{B}$ are thus shared between the DAUs. These weights are then applied to the respective locations (line 41) to yield the outputs of the DAUs per spatial location. As becomes apparent in line 41, the outputs are linear transformations (weighted sums) of the input and can be written as

$$\mathbf{a}_{l+1}(\mathbf{a}_l) = \mathbf{W}(\mathbf{a}_l)\mathbf{a}_l \quad , \tag{C.2}$$

with $\mathbf{a}_l \in \mathbb{R}^d$ the vectorised input to layer $l$ and $\mathbf{W} \in \mathbb{R}^{f \times d}$ and $f$ the number of filters (DAUs). The rows in matrix $\mathbf{W}$ correspond to exactly one filter (DAU) applied to exactly one patch $\mathbf{p}_{hw}$ and are non-zero only at those positions that correspond to this specific patch in the input.

---

[1]The best test accuracies per run are given by 79.16%, 79.04%, 78.86%, and 78.7% respectively.
[2]I.e., the ResNet weights are fixed in these experiments.

## C.3. Attribution methods

In section 5.2, we compare the model-inherent contribution maps of the CoDA-Nets to those of the following methods for importance attribution: the gradient of the class logits with respect to the input image [S2] (Grad), 'Input×Gradient' (IxG, cf. [S1]), GradCam [S13] (GCam), Integrated Gradients [S16] (IntG), DeepLIFT [S14], several occlusion sensitivities (Occ-K, with K the size of the occlusion patch) [S17], RISE [S10], and LIME [S11].

For RISE and LIME, we relied on the official implementations available at https://github.com/eclique/RISE and https://github.com/marcotcr/lime respectively. For RISE, we generated 6000 masks with parameters $s=6$ and $p_1=0.1$. For LIME, we evaluated on 256 samples per image and used the top 3 features for the attribution; for the segmentation, we also used the default parameters, namely 'quickshift' with $max\_dist = 200$, $ratio = 0.2$, and a kernel size of 4.

For Grad, GCam, IxG, IntG, DeepLIFT, and the occlusion sensitivities, we relied on the publicly available pytorch library 'captum' (https://github.com/pytorch/captum). GCam was used on the last activation map before global sum-pooling. The occlusion sensitivities were used with strides of 2 on CIFAR-10 and strides of 4 for TinyImagenet. Finally, for IntG we used 20 steps for the integral approximation.

## C.4. Evaluation metrics

In section 5.2, we evaluated the attribution methods against 2 quantitative metrics: **(1)** the adapted *pointing game* [S18] and **(2)** the prediction stability under removing the *least important pixels* as in [S15]. In section B, we further show results for removing the *most important pixels* first.

For **(1)**, we constructed 500 (250) $3\times 3$ multi-images for CIFAR-10 (TinyImagenet); for an example with $2\times 2$, see Fig. 7 in the main paper. In each of these multi-images, every class occurred at most once. As stated in section 5.2, we measured the fraction of positive contributions falling inside the correct mini-image. Further, the images were sorted according to their confidence for each of the classes. For every multi-image, a random set of classes was sampled. For each of the classes, we included the most confidently classified class image in the multi-image that had not been used yet in previous multi-images.

For **(2)**, we followed [S15] and successively replaced one (embedded) pixel at a time, replacing its embedding by a corresponding zero vector, until up to 25% of the image were removed. The pixels were removed in order, sorted by their assigned importance.

**Algorithm 1:** Implementation of a Convolutional Dynamic Alignment Layer

```python
1  from torch import nn
2  import torch.nn.functional as F
3
4  class DAUConv2d(nn.Module):
5
6      def __init__(self, in_channels, out_channels, rank, kernel_size, stride, padding, act_func):
7          # act_func: non-linearity for scaling the weights. E.g., L2 or SQ.
8          # out_channels: Number of convolutional DAUs for this layer.
9          # rank: Rank of the matrix AB.
10         # 'dim_reduction' applies matrix B.
11         self.dim_reduction = nn.Conv2d(in_channels, rank, kernel_size, stride, padding, bias=False)
12         # Total dimensionality of a single patch
13         self.patch_dim = in_channels * kernel_size * kernel_size
14         # 'weightings' applies matrix A and adds bias b.
15         self.weightings = nn.Conv2d(rank, out_channels * self.patch_dim, kernel_size=1, bias=True)
16         self.act_func = act_func
17         self.out_channels = out_channels
18         self.kernel_size = kernel_size
19         self.stride = stride
20         self.padding = padding
21
22     def forward(self, in_tensor):
23         # Project to lower dimensional representation, i.e., apply matrix B. This yields Bp for every patch p.
24         reduced = self.dim_reduction(in_tensor)
25         # Get new spatial size height h and width w
26         h, w = reduced.shape[-2:]
27         batch_size = in_tensor.shape[0]
28         # Apply matrix A and add bias b, yielding ABp + b for every patch p.
29         weights = self.weightings(reduced)
30         # Reshape for every location to size patch_dim×out_channels
31         weights = weights.view(batch_size, self.patch_dim, out_channels, h, w)
32         # Apply non-linearity to the weight vectors, yielding w(p) = g(ABp + b) as in eq. (1) for every patch p.
33         weights = self.act_func(weights, dim=1)
34         # Extract patches from the input to apply dynamic weights to patches.
35         patches = F.unfold(in_tensor, self.kernel_size, padding=self.padding, stride=self.stride)
36         # Reshape for applying weights.
37         patches = patches.view(batch_size, self.patch_dim, 1, h, w)
38         # Apply the weights to the patches.
39         # As can be seen, the output is just a weighted combination of the input, i.e., a linear transformation.
40         # The output can thus be written as o = W(x)x.
41         return (patches * weights).sum(1)
```

| Network | Input dimensions | Layer | Number of DAUs | Rank of AB | Kernel size | Stride |
|---|---|---|---|---|---|---|
| S-CoDA | $6 \times 32 \times 32$ | 1 | 16 | 32 | 3 | 1 |
| | | 2 | 16 | 32 | 3 | 1 |
| | | 3 | 32 | 64 | 3 | 2 |
| | | 4 | 32 | 64 | 3 | 1 |
| | | 5 | 32 | 64 | 3 | 1 |
| | | 6 | 64 | 64 | 3 | 2 |
| | | 7 | 64 | 64 | 3 | 1 |
| | | 8 | 64 | 64 | 3 | 1 |
| | | 9 | 10 | 64 | 1 | 1 |
| S-eCoDA | $6 \times 32 \times 32$ | 1 | 16 | 16 | 3 | 1 |
| | | 2 | 16 | 16 | 3 | 1 |
| | | 3 | 32 | 16 | 3 | 2 |
| | | 4 | 32 | 16 | 3 | 1 |
| | | 5 | 32 | 32 | 3 | 1 |
| | | 6 | 64 | 32 | 3 | 2 |
| | | 7 | 64 | 32 | 3 | 1 |
| | | 8 | 64 | 32 | 3 | 1 |
| | | 9 | 10 | 32 | 1 | 1 |
| L-CoDA | $6 \times 240 \times 240$ | 1 | 16 | 64 | 7 | 3 |
| | | 2 | 32 | 64 | 3 | 1 |
| | | 3 | 32 | 64 | 3 | 1 |
| | | 4 | 64 | 128 | 3 | 2 |
| | | 5 | 64 | 128 | 3 | 1 |
| | | 6 | 64 | 128 | 3 | 1 |
| | | 7 | 64 | 256 | 3 | 2 |
| | | 8 | 64 | 256 | 3 | 1 |
| | | 9 | 100 | 256 | 3 | 1 |
| XL-CoDA | $6 \times 64 \times 64$ | 1 | 16 | 64 | 5 | 1 |
| | | 2 | 32 | 64 | 3 | 1 |
| | | 3 | 32 | 128 | 3 | 2 |
| | | 4 | 64 | 128 | 3 | 1 |
| | | 5 | 64 | 128 | 3 | 1 |
| | | 6 | 64 | 256 | 3 | 2 |
| | | 7 | 64 | 256 | 3 | 1 |
| | | 8 | 64 | 256 | 3 | 1 |
| | | 9 | 200 | 256 | 3 | 2 |

Table C1: Architecture details for the results presented in Table 1 in the main paper.

## D. Low-rank matrix approximations

In the following, we first introduce the standard formulation of the low-rank matrix approximation problem and present the well-known solution via singular value decomposition. We then introduce an additional constraint to this standard formulation and show that at their optimum (maximal average output), the DAUs solve this constrained approximation problem.

**Low-rank approximation problem.** Given a data matrix[3] $\mathbf{M} \in \mathbb{R}^{m \times n}$, the low-rank approximation problem is typically formulated as $\min_{\mathbf{M}'}(||\mathbf{M} - \mathbf{M}'||_F)$ with $rank(\mathbf{M}') \leq r$ and $||\cdot||_F$ the Frobenius norm. The Frobenius norm can be written as a sum over the column differences $||\mathbf{\Delta}_i||_F^2 = ||\mathbf{m}_i||_2^2 + ||\mathbf{m}_i'||_2^2 - 2\mathbf{m}_i^T\mathbf{m}_i'$ with $\mathbf{m}_i, \mathbf{m}_i' \in \mathbb{R}^m$ the $i$-th columns of the matrices $\mathbf{M}$ and $\mathbf{M}'$ respectively. Note that since the $\mathbf{m}_i$ are fixed (they just correspond to the fixed input matrix $\mathbf{M}$), the optimisation objective is equivalent to

$$\min \sum_i ||||\mathbf{m}_i'||_2^2 - 2\mathbf{m}_i^T\mathbf{m}_i' = \tag{D.1}$$
$$\min \sum_i ||||\mathbf{m}_i'||_2^2 - 2||\mathbf{m}_i||||\mathbf{m}_i'||\cos(\alpha_i) \quad , \tag{D.2}$$

---

[3] In the context of our work, we can think of $\mathbf{M}$ as storing $n$ data samples of dimensionality $m$; e.g., when considering images, each column might correspond to one vectorised image from a dataset of $n$ images. The low-rank approximation problem aims to approximate this dataset with $r$ independent variables (dimensionality reduction).

25

with $\alpha_i$ the angle between $\mathbf{m}_i$ and $\mathbf{m}'_i$. Hence, optimising the Frobenius norm of the difference matrix finds a trade-off between aligning the directions (angles) of the columns of $\mathbf{M}$ and $\mathbf{M}'$ and approximating the correct norms of the original columns in $\mathbf{M}$. The optimal solution for this problem is given by singular value decomposition (SVD) [S4] of $\mathbf{M}$, and the optimal $\mathbf{M}'$ can be written as

$$\mathbf{M}' = \mathbf{U}_r \mathbf{U}_r^T \mathbf{M} = \mathbf{U}\boldsymbol{\Sigma}_r \mathbf{V}^T \quad , \tag{D.3}$$

where $\mathbf{U}_r$ is the matrix of left singular vectors of $\mathbf{M}$ up to the $r$-th vector[4]; the remaining vectors are replaced by zero-vectors. Note that the first formulation of the solution ($\mathbf{U}_r \mathbf{U}_r^T \mathbf{M}$) emphasises a property that we will encounter again for the DAUs. In particular, this formulation highlights the fact that the optimal matrix $\mathbf{M}'$ is given by reconstructing the individual data points from the $r$-dimensional eigenspace spanned by the eigenvectors of the data points; i.e., each column $\mathbf{m}'_i$ can be calculated as $\mathbf{U}_r \mathbf{U}_r^T \mathbf{m}_i$. The right hand side of the equation shows the conventional SVD-notation of the low-rank approximation solution, with $\mathbf{U} \in \mathbb{R}^{m \times m}, \boldsymbol{\Sigma}_r \in \mathbb{R}^{m \times n} \mathbf{V} \in \mathbb{R}^{n \times n}$ and $\boldsymbol{\Sigma}_r$ being a rectangular diagonal matrix with the first $r$ singular values as non-zero entries.

**Constrained low-rank approximations** Now we diverge from the standard low-rank approximation formulation and impose an additional constraint on $\mathbf{M}'$ (red box):

$$\min_{\mathbf{M}'}(||\mathbf{M} - \mathbf{M}'||_F) \,, \tag{D.4}$$

$$\text{s.t.} \quad rank(\mathbf{M}') \leq r \;\boxed{\wedge\; ||\mathbf{m}'_i||_2^2 = 1 \;\forall i} \,. \tag{D.5}$$

When combining eqs. (D.2) and (D.5), it becomes clear that the optimisation problem can now be formulated as

$$\max \sum_i \mathbf{m}_i^T \mathbf{m}'_i = \max \sum_i ||\mathbf{m}_i||_2 \cos\left(\angle(\mathbf{m}_i, \mathbf{m}'_i)\right) \,, \tag{D.6}$$

since the norm of $\mathbf{m}'_i$ is fixed and the addition of or multiplication with a fixed constant does not change the location of the optima.

As a corollary, we note that *any* matrix $\mathbf{M}'$ with a maximum rank $r$ that maximises eq. (D.6) gives a solution to the constrained approximation problem, *independent* of the norm of its columns: the columns can be normalised *post-hoc* to fulfill the new constraint, since rescaling does not change neither the rank of the matrix, nor the score of the objective function in eq. (D.6).

Further, we note that the unnormalised optimal matrix $\widetilde{\mathbf{M}}'$ can be factorised as

$$\widetilde{\mathbf{M}}' = \mathbf{A}\mathbf{B}\mathbf{M} \,, \tag{D.7}$$

$$\text{s.t.} \quad \mathbf{M}' = G(\widetilde{\mathbf{M}}') = G(\mathbf{A}\mathbf{B}\mathbf{M}) \tag{D.8}$$

with $\mathbf{A} \in \mathbb{R}^{m \times r}$ and $\mathbf{B} \in \mathbb{R}^{r \times n}$, similar to the solution to the unconstrained matrix approximation problem in eq. (D.3). Here, $G$ is a function that normalises the columns of a matrix to unit norm (as discussed in the previous paragraph, in order to fulfill the norm constraint, we can normalise the columns *post-hoc*).

Finally, this allows us to rewrite the objective function in eq. (D.6) as

$$\max \sum_i g(\mathbf{A}\mathbf{B}\mathbf{m}_i)\mathbf{m}_i = \max \sum_i \mathbf{w}(\mathbf{m}_i)\mathbf{m}_i \tag{D.9}$$

with $g(\mathbf{v})$ normalising its input vector to unit norm. Comparing this with eq. (1) in the main paper, we see that this is equivalent to maximising a DAU without bias term for maximal output over the $n$ columns in $\mathbf{M}$ and therefore, at their optimum, the DAUs solve the constrained low-rank matrix approximation problem in eq. (D.5). In particular, the DAUs achieve this by encoding common inputs in the eigenvectors of $\mathbf{A}\mathbf{B}$, which allows for an optimal angular reconstruction of the inputs, similar to reconstructing from the first PCA components in a PCA decomposition. The main difference to PCA is that PCA yields optimal reconstructions under the L2-norm, whereas the DAUs yield optimal angular reconstructions.

---

[4]A short proof of the equality to the right can be found at the end of this section.

**Proof of $\mathbf{U}_r\mathbf{U}_r^T\mathbf{M} = \mathbf{U}\boldsymbol{\Sigma}_r\mathbf{V}^T$.**

In order to see that this equality holds, we first write matrix $\mathbf{M}$ as

$$\mathbf{M} = \mathbf{U}\boldsymbol{\Sigma}\mathbf{V}^T \quad \text{(SVD-form)} \tag{D.10}$$

Now, we multiply both sides with $\mathbf{U}_r\mathbf{U}_r^T$ from the left:

$$\mathbf{U}_r\mathbf{U}_r^T\mathbf{M} = \mathbf{U}_r\mathbf{U}_r^T\mathbf{U}\boldsymbol{\Sigma}\mathbf{V}^T \tag{D.11}$$

$$\xrightarrow{\mathbf{U}_r^T\mathbf{U}=\mathbf{I}_r} \mathbf{U}_r\mathbf{U}_r^T\mathbf{M} = \mathbf{U}_r\mathbf{I}_r\boldsymbol{\Sigma}\mathbf{V}^T \tag{D.12}$$

$$\Rightarrow \mathbf{U}_r\mathbf{U}_r^T\mathbf{M} = \mathbf{U}\boldsymbol{\Sigma}_r\mathbf{V}^T \quad \square. \tag{D.13}$$

Here, we made use of the fact that $\mathbf{U}$ is an orthogonal matrix and therefore $\mathbf{U}^T\mathbf{U} = \mathbf{I}$; since we only use the first $r$ eigenvectors, we obtain the truncated identity matrix $\mathbf{I}_r$ when multiplying $\mathbf{U}_r^T\mathbf{U}$. Further, in the last line, we used that $\mathbf{U}\boldsymbol{\Sigma}_r = \mathbf{U}_r\boldsymbol{\Sigma}_r$.

## E. Comparison to capsule networks

In order to discuss the relationship to capsule networks, in section E.1 we will first rewrite the classical capsule formulation in 'Dynamic Routing Between Capsules' by Sabour et al. [S12] to mitigate the notational differences between their work and our work. In section E.2 we then show that while capsules and Dynamic Alignment Units share some computations, there are several important differences that we summarise in Table E1.

|               | **Classical capsules** | **Dynamic Alignment Units** |
|---------------|:---:|:---:|
| **Non-Lin.** $g$ | SQ | SQ, L2, ... |
| **Activations** | $g(\mathbf{V}\mathbf{x})$ | $g(\mathbf{A}\mathbf{B}\mathbf{x} + \mathbf{b})$ |
| **Routing**   | yes | no |
| **Low-rank**  | no | yes |
| **Output**    | CAP($\mathbf{x}$) | CAP($\mathbf{x}$)$^T\mathbf{x}$ |

Table E1: Comparison between capsules and Dynamic Alignment Units (DAUs). Importantly, DAUs produce a linear transformation of the input by multiplying the 'capsule activations' with the input (see 'Output') and allow for constraining the rank of the transformation. The dynamic weights in the DAUs can be seen as the activations of a capsule, s.t. $\mathbf{w}(\mathbf{x}) = $ CAP($\mathbf{x}$), see 'Output' in the table.

### E.1. Reformulating capsules

In this subsection, we will show that the classical capsule formulation (eq. (E.1)), in which input capsules 'vote' for the activations of an output capsule, can be written as a simple linear transformation $\mathbf{s} = \mathbf{V}\mathbf{x}$ if just one iteration of the dynamic routing algorithm is applied; here, $\mathbf{s}$ is a vector containing the activations of the output capsule, $\mathbf{V}$ stores the 'votes' of the input capsules to an output capsule, and $\mathbf{x}$ is a vector containing the activations of all input capsules. In the following, we will start from how capsules are formulated in [S12] and rewrite this formulation step by step.

In [S12] eq. (2), the authors calculate the activations $\mathbf{s}$ of a capsule[5] *before* any routing or non-linearity as

$$\mathbf{s} = \sum_i c_i \hat{\mathbf{u}}_i , \quad \hat{\mathbf{u}}_i = \mathbf{W}_i \mathbf{u}_i \quad . \tag{E.1}$$

Here, $\mathbf{u}_i$ is the capsule vector of capsule $i$ from the incoming layer and $\mathbf{W}_i$ a matrix which transforms the capsule activations to generate the votes $\hat{\mathbf{u}}_i$, i.e., the vote of the $i$-th incoming capsule to the output capsule in the current layer. Note that $c_i$

---
[5] As we only discuss a single output capsule, we omitted the subscript $j$ for the $j$-th output capsule for simplicity.

are the coefficients for the dynamic routing algorithm. If no routing is applied, they can be combined with $\mathbf{W}_i$ to yield $\widetilde{\mathbf{W}}_i = c_i \mathbf{W}_i$, which simplifies eq. (E.1) to

$$\mathbf{s} = \sum_i \hat{\mathbf{u}}_i , \quad \hat{\mathbf{u}}_i = \widetilde{\mathbf{W}}_i \mathbf{u}_i \quad . \tag{E.2}$$

Further, we note that the linear transformation of $\mathbf{u}_i$ can be represented as votes of the individual entries $t$ of $\mathbf{u}_i$:

$$\hat{\mathbf{u}}_i = \widetilde{\mathbf{W}}_i \mathbf{u}_i = \sum_t [\mathbf{u}_i]_t \left[ \widetilde{\mathbf{W}}_i^T \right]_t \quad . \tag{E.3}$$

Here, $[\mathbf{W}]_t$ denotes the $t$-th row in a matrix $\mathbf{W}$ (equivalently for a vector). Inserting this formulation of $\hat{\mathbf{u}}_i$ in eq. (E.2) yields

$$\mathbf{s} = \sum_i \sum_t [\mathbf{u}_i]_t \left[ \widetilde{\mathbf{W}}_i^T \right]_t \quad . \tag{E.4}$$

Hence, $\mathbf{s}$ can be represented as the result of votes from all neurons $u$ contained in any of the incoming capsules (note that we sum over all entries in all capsules). If we represent the activations of these neurons in a single vector $\mathbf{x}$, denote their activations by $x_u$, and their respective votes for the output capsule by $\mathbf{v}_u$ we can write eq. (E.4) as:

$$\mathbf{s} = \sum_i \sum_t \overbrace{[\mathbf{u}_i]_t}^{\hat{=} x_u} \overbrace{\left[ \widetilde{\mathbf{W}}_i^T \right]_t}^{\hat{=} \mathbf{v}_u} = \sum_u x_u \mathbf{v}_u \quad . \tag{E.5}$$

The sum on the right hand side in eq. (E.5), in turn, can be expressed as a simple matrix-vector multiplication, such that

$$\mathbf{s} = \sum_u x_u \mathbf{v}_u = \mathbf{V}\mathbf{x} \tag{E.6}$$

with $\mathbf{v}_u$ as the columns of $\mathbf{V}$. The result is, of course, trivial and only shows that the capsule activations (a weighted sum of the inputs, eq. (E.1)) are obtained as a linear transformation of the input if no dynamic routing is applied.

Finally, we note that subsequent to this linear transformation, the authors in [S12] apply the squashing function SQ (see eq. (2) in our main paper) to the output vector $\mathbf{s}$, which yields the final capsule output $\text{CAP}(\mathbf{x})$:

$$\text{CAP}(\mathbf{x}) = \text{SQ}(\mathbf{s}(\mathbf{x})) = \text{SQ}(\mathbf{V}\mathbf{x}) \quad . \tag{E.7}$$

### E.2. Differences to capsules

In the previous section we showed that the activation of a single capsule is computed as a linear transformation of all input activations, which is subsequently squashed (see eq. (E.7)). Comparing this with the computation of the DAU outputs (eq. (1) in our main paper), we see that this is equivalent to the dynamic weight computation if $\mathbf{V} = \mathbf{A}\mathbf{B}$ and $\mathbf{b} = \mathbf{0}$. As such, Dynamic Alignment Units and capsules are related. However, there are important differences, which we discuss in the following and summarise in Table E1.

First, in [S12], the squashed capsule activations $\text{CAP}(\mathbf{x})$ are used as input to the next layer (after potentially applying the dynamic routing algorithm first). Instead of forwarding the squashed activations directly, in our DAUs they are used to linearly transform the input. Second, by factorising the matrix $\mathbf{V}$ into two matrices $\mathbf{A}\mathbf{B}$, we are able to control the rank of the linear transformation. Third, when computing the weights in the DAUs we allow for an additional bias term, which in the context of capsules can be considered a 'default vote'. Fourth, we generalise the non-linearity in the DAUs to any non-linearity that only changes the norm. Lastly, we do not apply dynamic routing in the DAUs.


\end{document}